\documentclass[11pt]{article}

\usepackage[margin=1in]{geometry}
\usepackage{graphicx}
\usepackage{booktabs}
\usepackage{amsmath}
\usepackage{authblk}
\usepackage{hyperref}
\usepackage{natbib}
\usepackage{caption}
\usepackage{subcaption}
\usepackage{tabularx}

\graphicspath{{figures/}}

\title{Contextual Semantic Relevance Tracks fMRI BOLD Responses During Naturalistic Speech Comprehension}

\author[1]{Kun Sun}
\author[2]{Rong Wang}
\affil[1]{Department of Linguistics, Tongji University \thanks{kunsun@tongji.edu.cn}}
\affil[2]{Department of Computational Linguistics, Tuebingen University \thanks{rong.wang@uni-tuebingen.de}}

\date{}

\begin{document}
\maketitle

\begin{abstract}
Naturalistic language comprehension requires listeners to process both local probabilistic expectations and contextual semantic relations. This study tested whether contextual semantic relevance, a metric of how strongly a target word relates to its recent semantic context, is associated with fMRI BOLD responses independently of word surprisal and lexical, timing, acoustic, and prosodic controls. We analyzed two public datasets: \texttt{Alice} (23 participants hearing one literary narrative) and \texttt{Narratives} (47 participants contributing 185 runs from four shared stories). Original continuous ROI-level BOLD time series were analyzed using FIR/deconvolution, and run-standardized BOLD was analyzed with generalized additive mixed models (GAMMs). In Alice, semantic relevance was significant in all 12 ROIs in the FIR/deconvolution analyses, whereas surprisal survived FDR correction in none. In GAMMs, semantic relevance and surprisal were significant in 12/12 and 10/12 ROIs, respectively. In Narratives, FIR contrasts were significant in 17/20 ROIs for semantic relevance. In the GAMMs, semantic relevance and surprisal had FDR-significant smooth terms in 17/20 and 16/20 ROIs, respectively. Overall, semantic relevance showed robust associations with BOLD responses in both datasets, with a particularly strong advantage over surprisal in the timing-sensitive Alice FIR analysis. In Narratives, the two predictors showed comparable spatial prevalence, suggesting that contextual semantic fit and probabilistic expectation make partially distinct contributions that vary across datasets and analytical approaches. These findings suggest that contextual semantic fit and local probabilistic expectation make partially distinct, dataset-dependent contributions to hemodynamic responses during naturalistic language comprehension. Critically, the regionally heterogeneous direction of semantic relevance effects, with negative effects in posterior semantic regions and positive effects in frontal integration regions, suggests that semantic relevance is associated with functionally distinct neural processes rather than a single uniform neural mechanism.

\end{abstract}

\section{Introduction}

Understanding speech in natural contexts requires listeners to process a rapidly unfolding acoustic signal while building a coherent representation of events, entities, and relations over time. This process is incremental. Each incoming word is interpreted against prior context, but it can also update the listener's current interpretation. Naturalistic language paradigms are therefore useful because they preserve the temporal continuity, contextual richness, and semantic diversity of everyday comprehension \citep{hamilton2020revolution,huth2016semanticmaps}. They also make it possible to ask whether different computational dimensions of comprehension are expressed differently in neural data.

A central debate concerns the relation between prediction and integration. Information-theoretic accounts propose that the processing cost of a word is related to its negative log probability given the preceding context, or \emph{surprisal} \citep{hale2001earley,levy2008surprisal}. Prediction-based accounts further propose that comprehenders can use context to pre-activate likely upcoming linguistic input, although the strength, level, and necessity of prediction remain debated \citep{kuperberg2016prediction,pickering2018predicting}. Electrophysiological (EEG) evidence has often linked predictability and surprisal to N400-family responses, consistent with rapid lexical-semantic access or expectation-related processing \citep{kutas1984expectancy,kutas2011n400,delong2005preactivation,frank2015erp,lau2008n400}. Several fMRI studies have also reported surprisal-related BOLD effects, particularly in temporal and inferior frontal regions \citep{willems2016prediction,brennan2016abstract,shain2020fmri}, although these findings have not been consistently replicated across stimuli, language models, and analysis approaches.

However, naturalistic speech comprehension also requires semantic integration. Incoming words are related to recently processed words, combined with semantic memory, and incorporated into an evolving interpretation of the narrative. Neurobiological models of language emphasize that comprehension depends on multiple interacting operations, including memory retrieval, unification or integration, prediction, and control \citep{hagoort2013muc,lau2008n400}. Retrieval-based accounts further highlight the role of memory access mechanisms in sentence processing, whereby each incoming word triggers cue-based retrieval from a decaying memory representation \citep{lewis2005activation}. In addition, semantic cognition is supported by distributed temporal, inferior parietal, frontal, and medial systems rather than by a single classical language area \citep{binder2011semanticmemory,lambonralph2017semantic,seghier2013angular}. Naturalistic fMRI studies are especially relevant here because they show that semantic information during story comprehension is represented across broad cortical systems \citep{wehbe2014harrypotter,huth2016semanticmaps,jain2018context}.

These theoretical debates are closely tied to measurement timescale. Surprisal may index a fast local prediction-error-like signal, which is well suited to EEG and eye-movement measures. By contrast, fMRI measures a delayed and temporally smoothed BOLD response \citep{boynton1996linear,logothetis2003bold}. In continuous speech, words arrive rapidly and adjacent hemodynamic responses overlap, potentially making brief word-level prediction-error effects difficult to isolate. At the same time, higher-order cortical regions integrate information over longer temporal windows during naturalistic comprehension \citep{hasson2008hierarchy,lerner2011temporal,hasson2015hierarchical}. BOLD may therefore be more sensitive to contextual semantic variables that unfold over several seconds than to brief word-level fluctuations in probabilistic unexpectedness.

To understand the role of contextual semantic information, the current study examines a computational metric that we define as \emph{(contextual) semantic relevance}. Operationally, semantic relevance is computed as a distance-weighted combination of cosine similarities between a target word's embedding and those of its three preceding context words, adjusted for pairwise similarity already present among the context words (see Section~\ref{sec:semrel} for the detailed computation). Rather than exploring how unexpected a word is, semantic relevance measures how strongly a target word relates to, fits with, or contributes to the recent semantic context. A word can be locally unexpected but still semantically appropriate, and a predictable word can contribute little additional semantic structure. This distinction motivates the hypothesis that surprisal and semantic relevance are not simply alternative labels for the same process. Instead, they may index partially different computations and timescales. Surprisal reflects local probabilistic expectation, whereas semantic relevance reflects contextual semantic fit or integration.

The present study examines this account using two public naturalistic speech-comprehension fMRI datasets with different statistical methods. The Alice dataset provides strong group alignment because all participants listened to the same short literary narrative \citep{bhattasali2020alice}. The Narratives dataset provides a larger independent sample in which participants listened to four shared stories with verified word-level timestamps \citep{nastase2021narratives}. If surprisal and semantic relevance reflect partially distinct computations with different temporal characteristics, they should exhibit distinguishable neural signatures. Semantic relevance may produce delayed, spatially distributed BOLD associations consistent with sustained contextual processing, whereas surprisal may be more difficult to isolate in the temporally smoothed hemodynamic signal. Testing this hypothesis requires naturalistic speech datasets, complementary analytical frameworks with different temporal sensitivities, and rigorous control of lexical, acoustic, and prosodic confounds. 

The current study addresses three questions: (1) Does semantic relevance relate to BOLD after lexical, timing, acoustic, prosodic, participant, run, and story-related variables are controlled where available, and does it show an HRF-consistent delayed profile? (2) Are semantic relevance effects more spatially prevalent than surprisal effects across datasets and modeling frameworks? (3) Do the two metrics show distinguishable ROI-level patterns consistent with partially different computations?

\section{Materials and Methods}

\subsection{Datasets}

\paragraph{Alice fMRI dataset.}
The Alice dataset (abbreviated as \texttt{Alice}) contains fMRI data from participants listening to the first chapter of \emph{Alice's Adventures in Wonderland} \citep{bhattasali2020alice}. The fMRI sample includes 26 participants who listened passively to the same 12.4-minute audiobook stimulus. The stimulus contains 2,129 words and was acquired with TR (repetition time) = 2 s, yielding 372 time points. This dataset provides a compact and highly aligned naturalistic design: all participants heard the same literary narrative, the same word sequence, and the same discourse progression. This makes \texttt{Alice} especially useful for group-level ROI analyses and for estimating shared delayed response profiles with FIR/deconvolution. In the present study, the Alice dataset serves as the higher-participant, tightly aligned test case for whether semantic relevance is associated with BOLD responses during naturalistic story comprehension.

\paragraph{Narratives fMRI dataset.}
The Narratives dataset (abbreviated as \texttt{Narratives}) is a public collection of naturalistic story-listening fMRI datasets comprising 345 participants, 891 functional scans, and 27 spoken stories. It provides BIDS-formatted imaging data together with preprocessed derivatives, stimulus materials, and time-aligned word- and phoneme-level transcripts \citep{nastase2021narratives}. The present analysis used the subset of 47 participants who contributed data for the four shared English stories \texttt{black}, \texttt{bronx}, \texttt{forgot}, and \texttt{piemanpni}. After file-integrity and alignment checks, 185 official \texttt{afni-nosmooth} clean BOLD runs were available. Of the 47 participants, 45 contributed all four stories, one contributed three stories, and one contributed two stories. Functional MRI data were acquired with a TR of 1.5 s. Across the four stimuli, 5,999 of 6,050 transcript word instances were aligned to Gentle word onsets and retained for predictor construction.

The official derivatives were produced with fMRIPrep 20.0.5 and subsequently cleaned with AFNI \texttt{3dTproject}. The cleaning model included six motion parameters, the first five principal components from cerebrospinal-fluid and white-matter signals, a 128-s high-pass cosine basis, and first- and second-order polynomial trends \citep{nastase2021narratives}. We analyzed these unsmoothed clean residuals without an additional temporal shift and retained all acquired TRs. For each run and ROI, the primary response was the clean residual standardized to zero mean and unit variance within run, and analyses in the original residual units were used as a sensitivity check.

%Twenty bilateral anatomical ROIs were defined from the Harvard--Oxford maximum-probability atlas (25\% threshold, 2-mm template). Atlas labels were combined across hemispheres and resampled with nearest-neighbor interpolation to each BOLD grid; the BOLD images themselves were not resized. The ROIs were AG, CG, FOC, HG, IC, IFG, MTG, PT, SMG, STG, TP, ITG, FUS, PP, PHG, HPC, PCUN, PCC, vmPFC, and dmPFC/SFG. Each ROI contributed 83,558 TR-level observations across the 185 runs.

\paragraph{Complementary role of the two datasets.}
Alice and Narratives provide complementary tests. Alice offers tight alignment to one stimulus and supports comparison with the original analysis. Narratives provides a larger sample, four shared stories, official nuisance-cleaned derivatives, true word onsets, and explicit run and story structure. Convergence would therefore indicate that an association is not specific to one story or preprocessing pipeline; divergence is also informative because the datasets differ in stimulus set, response construction, and the operationalization of semantic relevance.

\subsection{Computational Metrics}

We compared two word-level computational metrics: word surprisal and semantic relevance. The central theoretical contrast is that surprisal measures local probabilistic unexpectedness, whereas semantic relevance measures contextual semantic fit.

\paragraph{Surprisal.}
Surprisal was treated as a metric of local probabilistic unexpectedness. For a target word $w_t$, surprisal is defined as:

\begin{equation}
\mathrm{Surprisal}(w_t) = -\log P(w_t \mid w_{<t}).
\end{equation}

In both datasets, word-level surprisal was computed with \texttt{GPT-2}. In the Alice dataset, we computed word-level surprisal using \texttt{GPT-2} rather than relying on the available surprisal annotations. In both datasets, surprisal was likewise derived primarily from \texttt{GPT-2}, with \texttt{GPT-Neo} used as a robustness check. Subword negative log probabilities were summed to obtain one surprisal value per orthographic word, in natural-log units. Higher values therefore indicate words that are less probable given the preceding text. %Surprisal was interpreted as a measure of local probabilistic unexpectedness: higher values indicate words that are less expected given the preceding context.

\paragraph{Semantic relevance.}
\label{sec:semrel}
Semantic relevance quantifies how strongly a target word is semantically connected to its recent local context. Unlike surprisal, which estimates how unexpected a word is, semantic relevance estimates the degree of semantic fit between the target word and the words that have been processed. The metric was motivated by the assumption that local contextual information is not equally available during incremental comprehension. Recent words tend to have stronger influence on current processing than more distant words \citep{cowan2001magical,oberauer2002access,lewis2005activation,christiansen2016now}. This local-context view is also consistent with EEG and eye-tracking evidence that short-range semantic context predicts neural and reading-time responses during naturalistic comprehension \citep{frank2017similarity,broderick2018semantic,sun2026eeg,sun2023interpretable,sun2024visual,sun2024attention1,sun2022semantic,sun2026speech,sun2026attention}.

The general semantic relevance algorithm combines two sources of information: target-context semantic fit and context-context semantic coherence. For a target word $w_t$, we represent the target and its three preceding context words, $w_{t-3}$, $w_{t-2}$, and $w_{t-1}$, using embedding vectors. In static embedding implementations, these vectors were derived from \texttt{Model2vec} word embeddings \citep{minishlab2024model2vec}. We first compute the cosine similarity between the target vector and each context-word vector. These target-context similarities estimate how strongly the target word fits the recent semantic context. To reflect graded contextual availability, the three preceding words receive distance-based weights, with closer words receiving larger weights.

The metric also includes pairwise semantic similarities among the three preceding context words. These context-context similarities capture the local semantic coherence of the material already processed before the target word is encountered. In the present metric, target-context similarity was treated as the direct semantic fit between the incoming word and the recent context, whereas context-context similarity was treated as the background coherence already present in the preceding context. Thus, semantic relevance combines two sources of information: the semantic fit between the target word and the recent context, and the semantic coherence within that context.

Formally, semantic relevance for target word $w_t$ is:

\begin{equation}
	\mathrm{SemRel}(w_t) =
	\sum_{k \in \{t-3,t-2,t-1\}} \alpha_k \cdot \mathrm{Sim}(v_t, v_k)
	+
	\sum_{(k,l) \in \mathcal{P}} \beta \cdot \mathrm{Sim}(v_k, v_l),
\end{equation}

where $v_i$ denotes the embedding vector for word $w_i$, $\mathcal{P}=\{(t-3,t-2),(t-3,t-1),(t-2,t-1)\}$, and $\beta$ is the fixed weight assigned to pairwise context-context similarity. The second term captures the local semantic coherence already present among the preceding context words. Cosine similarity is defined as:

\begin{equation}
	\mathrm{Sim}(v_i,v_j) = \frac{v_i \cdot v_j}{\lVert v_i\rVert \lVert v_j\rVert}.
\end{equation}

\begin{figure}[!h]
	\centering
	\includegraphics[width=0.82\textwidth]{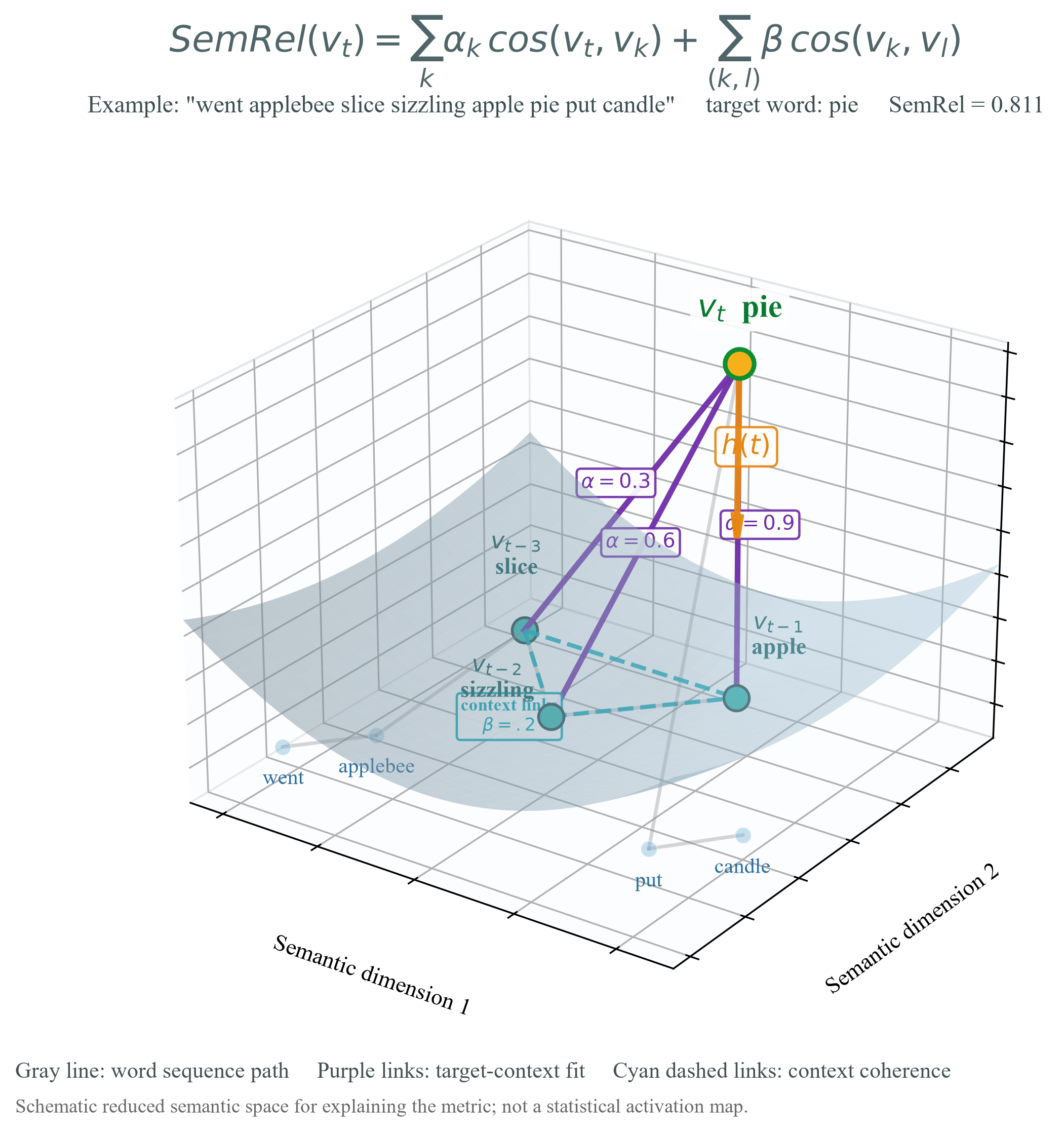}
	%  \caption{Alice FIR/deconvolution results for the dataset-provided semantic similarity metric. The figure summarizes ROI-level significance and delayed lag profiles in original BOLD. The semantic similarity metric was significant in all 12 ROIs, whereas surprisal was not significant in any ROI after FDR correction.}
	%  \caption{
		%Schematic illustration of the semantic relevance metric in a reduced semantic-vector space. The example is ``... went applebee slice sizzling apple pie put candle'', with ``pie'' as the target word. The gray line shows the word sequence path. The three immediately preceding words, ``slice'', ``sizzling'', and ``apple'', form the local context. Purple links indicate target-context semantic fit, computed as cosine similarity between the target-word vector and each context-word vector, weighted by recency weights $\alpha_k$. Cyan dashed links indicate the context-coherence correction, computed from pairwise similarities among the local context words and weighted by $\beta$. The orange arrow schematically indicates that the word-level semantic relevance predictor is later mapped to BOLD responses through hemodynamic modeling. The figure is intended as a conceptual visualization of the metric and is not a statistical activation map.
		%}
	\caption{Schematic illustration of the semantic relevance metric in a reduced semantic-vector space. The example is ``... went applebee slice sizzling apple pie put candle'', with ``pie'' as the target word. The gray line shows the word sequence path. The three immediately preceding words, ``slice'', ``sizzling'', and ``apple'', form the local context. Purple links indicate target-context semantic fit, computed as cosine similarity between the target-word vector and each context-word vector, weighted by recency weights $\alpha_k$. Cyan dashed links indicate the context-coherence component, computed from pairwise similarities among the local context words and weighted by $\beta$. The orange arrow schematically indicates that the word-level semantic relevance predictor is later mapped to BOLD responses through hemodynamic modeling. The figure is intended as a conceptual visualization of the metric and is not a statistical activation map.
	}
	\label{fig:example}
\end{figure}

The distance-based target-context weights were generated from a linear recency-decay function. Motivated by time-based working-memory decay models, we approximated recency-based contextual availability as a word-distance decay function \citep{oberauer2011modeling,gauvrit2016mathematical}. During online comprehension, recently encountered words are assumed to remain more accessible in working memory and to exert stronger influence on the interpretation of the target word, whereas more distant words become less available as memory traces decay or are displaced by intervening material \citep{cowan2001magical,oberauer2002access,lewis2005activation,christiansen2016now}. We therefore assigned larger weights to more recent context words using the following function:

\begin{equation}
	\alpha_{t-d} = \alpha_{\max} \cdot \frac{m-d+1}{m},
\end{equation}

where $d$ denotes the distance between the target word and the context word, $m$ is the size of the local context window, and $\alpha_{\max}$ is the maximum weight assigned to the immediately preceding word. In the present study, we used a three-word context window ($m=3$) and set $\alpha_{\max}=0.9$, yielding $\alpha_{t-1}=0.9$, $\alpha_{t-2}=0.6$, and $\alpha_{t-3}=0.3$. The context-context weight was fixed at $\beta=0.2$, lower than the target-context weights, because this term captures background semantic coherence among the preceding words rather than the direct fit of the target word itself. All weights and the window size were fixed before any fMRI modeling and should not be interpreted as parameters optimized on brain data. Alternative weighting functions, such as exponential decay or equal weighting, and alternative window sizes would be useful targets for future confirmatory analyses. A higher semantic relevance value indicates stronger semantic fit between the target word and its recent context, together with stronger semantic coherence within that context. The computation of semantic relevance is illustrated in Figure~\ref{fig:example}.

\paragraph{Control predictors.}
To evaluate the effects of the main predictors of interest (surprisal and semantic relevance), we included a comprehensive set of lexical, timing, acoustic, and prosodic control variables in all statistical models. In \texttt{Alice}, these controls comprised log word frequency (recomputed as English Zipf frequency using \texttt{wordfreq} \citep{robyn_speer_2022_7199437}), word length, word duration, root mean square (RMS) acoustic amplitude, and prosodic-break strength. Word duration was defined as the interval between word onset and offset. RMS amplitude, reflecting the average acoustic energy of each spoken word, was calculated over the word's duration from the provided audio stimuli. In \texttt{Narratives}, the control variables included log word frequency, word length, word duration, and word rate, defined as the number of word onsets within each TR. The official \texttt{Narratives} residual BOLD data had already been denoised to remove motion-related variance, tissue-derived nuisance components, low-frequency drift, and polynomial temporal trends. In addition, participant, run, story, and within-story temporal structure were modeled as described in the GAMM analysis section.

\subsection{FIR/deconvolution Analysis of Original BOLD}

To test hemodynamic response profiles, we modeled original continuous BOLD time series using finite impulse response (FIR) deconvolution, which estimates separate response amplitudes at successive post-stimulus delays without assuming a fixed hemodynamic response shape \citep{dale1999optimal,glover1999deconvolution,henson2001choice,lindquist2009modeling}. FIR/deconvolution preserves the continuous time series and estimates separate effects at multiple post-word-onset delays. This makes it possible to ask whether a predictor has the delayed temporal profile expected of a hemodynamic response.

For each predictor, lagged regressors were constructed across post-word-onset delays. Alice FIR models tested semantic relevance and surprisal while controlling for FIR-lagged log (word frequency), word length, word duration, slow time, and subject effects. Grouped FIR tests assessed whether each predictor explained BOLD variance across the full lag profile.  In Alice, the FIR regressors spanned post-word-onset delays from 0 to 16~s in 2-s steps, matching the TR and yielding nine lag regressors per predictor. The dataset documentation reports that 26 participants remained
after the original exclusion of one participant for excessive head motion and two for poor behavioral performance. The present FIR analysis included 23 participants because usable preprocessed BOLD images were available for only 23 participants in the released derivatives. This standard FIR/deconvolution approach was appropriate for Alice because all participants listened to the same short narrative, providing strong temporal alignment across subjects.

For \texttt{Narratives}, event regressors were constructed from the true Gentle word onsets after adding the BIDS stimulus-onset offset for each run. Semantic relevance, surprisal, log frequency, word length, duration, and word rate were represented at lags from 0 to 15 s in 1.5-s steps. Models were estimated separately for each participant and ROI, preventing participants with more runs from receiving greater inferential weight. The prespecified summary contrast used the 4.5--12-s lags, weighted by nonnegative values from a canonical SPM-like HRF and normalized to sum to one. Group inference used 100,000 participant-level random sign flips. Two-sided \textit{p}-values were computed separately for semantic relevance and surprisal and then corrected across the 20 ROIs using the Benjamini--Hochberg false discovery rate (FDR) procedure to control the expected proportion of false positives among significant findings. Directional estimates are reported descriptively rather than used to redefine the test after observing the data. The primary response was run-standardized clean BOLD; the same models were repeated in official residual units and in the 45 participants with all four stories.

\subsection{Transformed BOLD and HRF-convolved Predictors}

We used transformed BOLD data for complementary statistical analyses. In the present study, transformed BOLD refers to derived ROI-level response tables in which the BOLD signal was aligned with word-level or TR-level linguistic predictors and represented as an analyzable response for each ROI, participant, and stimulus time point. As these tables depend on temporal alignment, interpolation, HRF convolution or shifting, and the mapping between words and TRs, they were not used as the primary basis for timing-sensitive inference.

For \texttt{Alice}, we retained the 12 ROIs defined in the established dataset-specific analysis pipeline, which encompass regions associated with language, speech, and emotion processing: angular gyrus (AG), cingulate gyrus (CG), frontal opercular cortex (FOC), Heschl's gyrus (HG), insular cortex (IC), inferior frontal gyrus (IFG), middle temporal gyrus (MTG), planum temporale (PT), subcallosal cortex (SG), supramarginal gyrus (SMG), superior temporal gyrus (STG), and ventromedial prefrontal cortex (vmPFC). The transformed BOLD dataset was organized at the word-by-ROI level, with each observation containing the linguistic predictors, lexical and acoustic control variables, participant identifier, and transformed BOLD response. The complete-case GAMM analysis included data from 23 of the 26 participants, yielding 7,866 observations per ROI.

Compared with Alice, \texttt{Narratives} includes a more diverse set of stories, more participants, and a larger overall dataset. Accordingly, we extracted 20 bilateral ROIs using the Harvard–Oxford probabilistic atlas implemented in FSL \citep{smith2004advances,jenkinson2012fsl}.
This set included regions involved in core auditory and language processing: AG, FOC, HG, IC, IFG, MTG, PT, SMG, STG, temporal pole (TP), inferior temporal gyrus (ITG), fusiform gyrus (FUS), and planum polare (PP), together with regions implicated in narrative integration, episodic memory, and the default-mode network: CG, parahippocampal gyrus (PHG), hippocampus (HPC), precuneus (PCUN), posterior cingulate cortex (PCC), vmPFC, and dorsomedial prefrontal/superior frontal cortex (dmPFC/SFG). ROI definitions were fixed before statistical testing and were not selected according to the significance of semantic relevance or surprisal effects.

The broader Narratives ROI set was possible because this dataset provided standardized whole-brain data from 47 participants listening to all four shared stories, thereby offering substantially greater participant-level and stimulus-level coverage than Alice. The expansion from 12 to 20 ROIs was intended to sample both the core language network and the distributed memory and situation-model systems engaged during extended narrative comprehension. It should therefore be understood as a theoretically motivated expansion of anatomical coverage, rather than as an attempt to increase the number of significant results. Analyses comparing the two datasets directly were restricted to anatomically corresponding ROIs.

For Narratives, each word-level predictor was placed at its true acoustic onset, convolved with the same canonical HRF, and sampled at TR resolution. The dependent variable was the official clean ROI residual standardized within run. No FIR-motivated deletion of initial TRs was applied to the GAMM data. This transformed-BOLD strategy is complementary to FIR/deconvolution but has important limitations. Transformed-BOLD models are useful for characterizing conditional partial-effect shapes, including possible nonlinear associations, but they do not recover the complete hemodynamic response profile. Moreover, word-level tables can inflate apparent precision when multiple word observations inherit information from the same TR-level BOLD sample. We therefore treated transformed-BOLD GAMMs as complementary effect-shape analyses, whereas FIR/deconvolution of the original continuous BOLD signal provided the primary timing-sensitive test of delayed hemodynamic effects.

\subsection{GAMM Analysis of Transformed BOLD}

We modeled transformed BOLD data using generalized additive mixed models (GAMMs) \citep{wood2017generalized}. GAMMs are useful because they allow smooth nonlinear effects of continuous predictors while also including random-effect terms for participants. However, because the transformed-BOLD tables are not the primary unit for hemodynamic timing inference, these GAMMs were interpreted as supplementary effect-shape models. For Alice, each ROI was fit using:

\begin{equation}
\begin{split}
\text{bold} \sim &\ te(\text{log\_freq}, \text{wordlen}, k=5) + s(\text{word duration}, k=5) \\
&+ s(\text{RMS}, k=5) + s(\text{pros}, k=5) + s(\text{surprisal}, k=5) \\
&+ s(\text{semrel}, k=5) + s(\text{subj}, \text{bs}=\text{``re''}).
\end{split}
\end{equation}

%Semrel is the abbreviation of ``semantic relevance''.

The GAMM was implemented using the \texttt{mgcv} package in \texttt{R}. In the model specification, the response variable (\texttt{bold}) denotes the transformed BOLD response. The primary predictors of interest were surprisal and semantic relevance (\texttt{semrel}), whereas log word frequency (\texttt{log\_freq}), word length (\texttt{wordlen}), word duration, root mean square (RMS) acoustic amplitude, and prosodic-break strength (\texttt{pros}) were included as control variables. The function \texttt{s()} specifies a univariate penalized regression spline to capture potentially nonlinear relationships between each predictor and the BOLD response. The argument \texttt{k = 5} sets the basis dimension, limiting the maximum complexity of each smooth while allowing the effective degrees of freedom to be determined by penalized smoothing during model fitting. The term \texttt{s(subj, bs = "re")} specifies participant as a random effect, where \texttt{bs = "re"} indicates a random-effects basis that models participant-specific intercepts. The function \texttt{te()} specifies a tensor-product smooth for the joint effects of log word frequency and word length. Because these lexical variables are moderately correlated, modeling them jointly with a tensor-product smooth allows flexible estimation of their combined nonlinear effects while reducing potential collinearity between separate smooth terms.

These models were fit with \texttt{bam(method = "fREML", discrete = TRUE)}. For Narratives, HRF-convolved TR-level predictors were used:

\begin{equation}
\begin{split}
\text{bold}_{z} \sim &\ te(\text{log\_freq}_{HRF}, \text{word length}_{HRF}, k=5) \\
&+ s(\text{duration}_{HRF}, k=5) + s(\text{word rate}_{HRF}, k=5) \\
&+ s(\text{TR},\text{story},\text{bs}=\text{``fs''},k=10) \\
&+ s(\text{participant},\text{bs}=\text{``re''}) + s(\text{run},\text{bs}=\text{``re''}) \\
&+ s(\text{semrel}_{HRF}, k=5) + s(\text{surprisal}_{HRF}, k=5).
\end{split}
\end{equation}

The GAMMs fittings for Narratives used \texttt{bam(method = "ML")}, reproducing the previously selected GAMM specification while extending it to all 20 ROIs. For Narratives, word-level values were placed at their verified scanner-aligned
onsets, accumulated within TR, convolved with an SPM-like double-gamma HRF,
and standardized after sampling at the 1.5-s TR. As a result, the suffix
\texttt{\_hrf} denotes preprocessing of the predictor before GAMM fitting, and it
does not indicate that the GAMM applied an additional HRF transformation.
The same conceptual operation had already been incorporated into the legacy
Alice predictor table, although it was not indicated in the Alice column
names.

%Predictor evidence was assessed in two ways: approximate smooth-term tests from the full model and maximum-likelihood nested deletion tests comparing the full model with a model omitting one target smooth. Because residual temporal autocorrelation can make these tests anti-conservative, a run-aware sensitivity analysis used fREML, an AR(1) error term initialized at each run boundary, story-specific time trends, and participant-level random slopes for the target predictors. The smooth terms estimate conditional partial associations with transformed BOLD, not held-out predictive gain.

\paragraph{Participant-specific random-slope sensitivity analysis.}

To examine whether the associations of the main predictors varied across
listeners, we additionally fitted a participant-heterogeneity sensitivity
model containing participant-specific random slopes for semantic relevance
and surprisal based on the original GAMM fittings for both datasets:

\begin{equation}
	s(\mathrm{participant}_{i},
	by=\mathrm{semrel}^{\mathrm{HRF}}_{st},
	bs=\text{``re''})
	+
	s(\mathrm{participant}_{i},
	by=\mathrm{surprisal}^{\mathrm{HRF}}_{st},
	bs=\text{``re''}).
\end{equation}

These terms allow each participant to deviate linearly from the corresponding
population-level predictor effect, while
$s(\mathrm{semrel}^{\mathrm{HRF}},k=5)$ and
$s(\mathrm{surprisal}^{\mathrm{HRF}},k=5)$ estimate the population-level
potentially nonlinear associations. The HRF-convolved predictors were
standardized before model fitting. A participant random intercept was retained
to represent differences in mean BOLD level, and the random slopes represented
between-participant heterogeneity in predictor sensitivity. The random-slope terms were included to reduce the assumption that all
participants shared identical predictor--BOLD associations. They should not
be interpreted as separate significance tests or predictive-performance
estimates for individual participants. %Participant-level generalization was evaluated more directly by the participant-level FIR analysis.

\paragraph{Complementary role of GAMM and FIR/deconvolution.}
The two statistical approaches answer related but different questions. GAMMs on transformed BOLD investigates whether a predictor is associated with BOLD amplitude and what the partial-effect curve looks like after controlling for other variables. FIR/deconvolution explores when the effect appears in the original BOLD time series and whether it follows a plausible delayed HRF profile. In this way, GAMM provides an interpretable but supplementary effect-shape analysis, whereas FIR/deconvolution provides timing-sensitive evidence about delayed BOLD responses. Since the transformed-BOLD and FIR analyses can differ in scaling, centering, and temporal alignment, apparent agreement or disagreement between their effect directions should be interpreted cautiously.

\paragraph{Multiple-comparison correction and GAMM inference}

ROI-level \textit{p}-values were corrected using the Benjamini--Hochberg false-discovery-rate (BH--FDR) procedure within predefined test families. For Alice, corrections were performed separately for semantic relevance and surprisal across the 12 ROIs and separately for the FIR and GAMM analyses. For Narratives, corrections were performed separately for the two predictors across the 20 ROIs within each analysis: participant-level FIR/deconvolution, GAMM smooth-term testing, and valid nested model-deletion comparisons. Unless otherwise stated, statistical
significance was defined as $q<.05$.

For the GAMMs, inference on each target predictor was based on the
approximate smooth-term test provided by \texttt{mgcv}. We report the
estimated effective degrees of freedom (EDF), approximate $F$ statistic,
uncorrected \textit{p} value, and BH--FDR-corrected \textit{q} value. An EDF
near 1 indicates that the estimated association is approximately linear,
whereas larger values indicate greater fitted nonlinearity. These tests
evaluate evidence for the corresponding smooth conditional on the other
terms in the fitted model; they should not be interpreted as direct measures
of out-of-sample predictive improvement.

Partial-effect plots show the centered estimated smooth for the target
predictor with pointwise 95\% confidence intervals. They are presented to
illustrate the estimated functional form rather than to determine
significance. Statistical conclusions were based on the FDR-corrected
smooth-term tests or valid nested model comparisons, not on whether a
confidence band crossed zero at particular predictor values.

\section{Results}

\subsection{Analysis overview}

The Alice analysis used 12 predefined ROIs: AG, CG, FOC, HG, IC, IFG, MTG, PT, SG, SMG, STG, and vmPFC. The Narratives analysis used 20 bilateral ROIs spanning auditory, temporal, inferior frontal, parietal, medial temporal, and default-mode systems: AG, CG, FOC, HG, IC, IFG, MTG, PT, SMG, STG, TP, ITG, FUS, PP, PHG, HPC, PCUN, PCC, vmPFC, and dmPFC/SFG. %The dataset-specific ROI sets and broad functional associations are summarized in Table~\ref{tab:roi_functions}.
These ROIs collectively sample distributed processes involved in story listening. HG, STG, MTG, PT, and ITG contribute to auditory, phonological, lexical, and semantic processing; IFG, FOC, IC, and dmPFC/SFG contribute to selection, control, and integration; AG, SMG, PP, and PCUN contribute to multimodal and narrative integration; and HPC, PHG, PCC, vmPFC, and TP contribute to memory, contextual association, and situation-model construction. The commonly associated functions of the individual ROIs are summarized in Table~\ref{tab:roi_functions} in Appendix~A.

Moreover, the word-level Pearson correlation analyses indicated that semantic relevance and surprisal were only weakly related in both datasets. The correlation was weakly negative in Alice ($r=-.19$) and closer to zero in Narratives ($r=-.07$). The expanded heatmaps also included the dataset-specific control variables: word length, log frequency, word duration, RMS, and prosodic break in Alice, and word length, log frequency, word duration, and word rate in Narratives. The strongest correlations occurred primarily among lexical and timing controls, particularly word length, word duration, and log frequency. These results indicate that the two target predictors capture largely nonredundant word-level properties and that their BOLD associations are unlikely to arise simply from direct collinearity with one another. However, a weak zero-order correlation does not by itself demonstrate independent neural contributions or establish that one predictor is stronger, and those conclusions require the conditional GAMM, nested-model, or held-out prediction analyses. The complete correlation heatmaps are presented in Appendix~A (Figure~\ref{fig:corr}).

\subsection{Results of the Alice Dataset}

\subsubsection{Semantic relevance for FIR/deconvolution effects}

\begin{figure}[t]
    \centering
    \includegraphics[width=\textwidth]{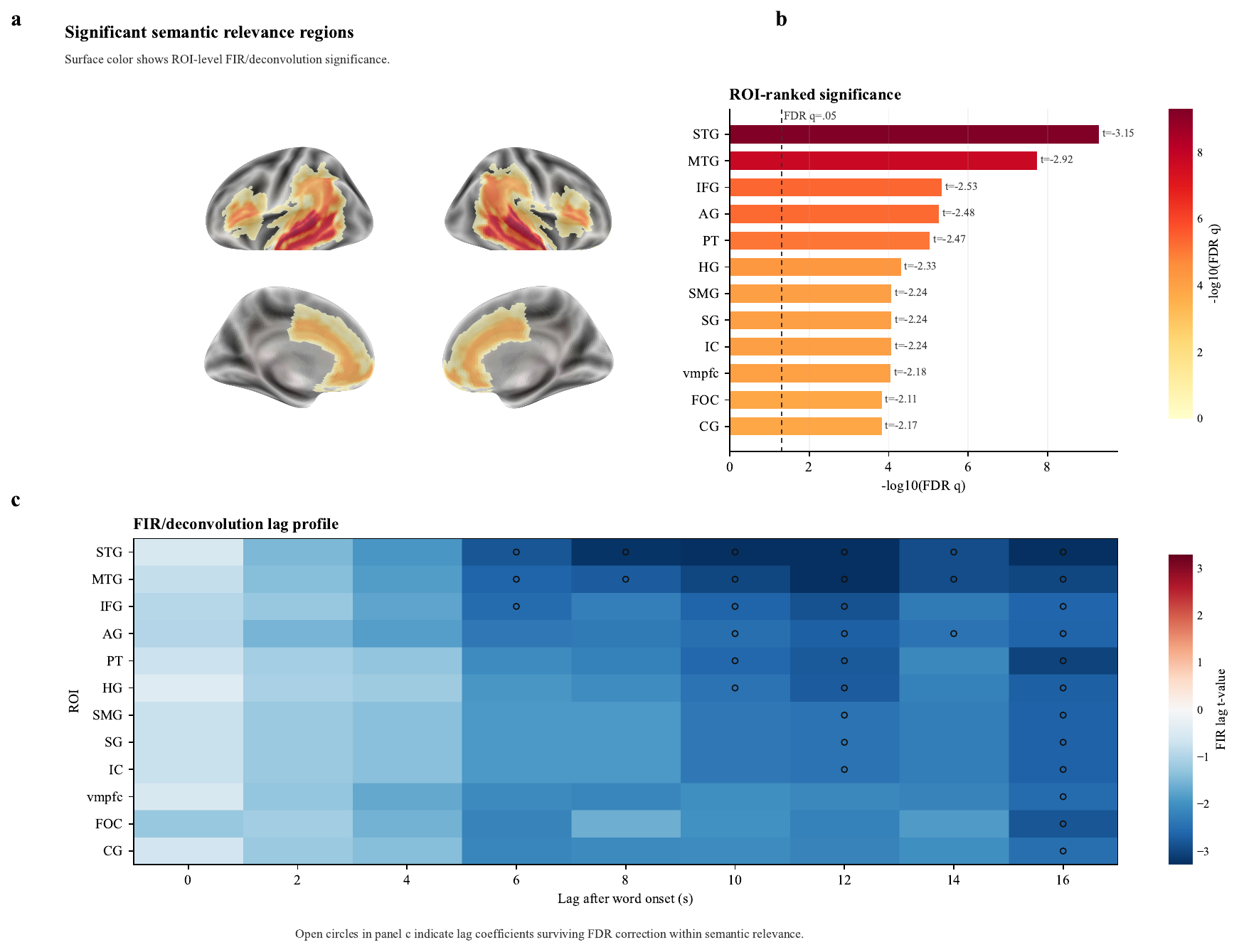}
    \caption{Alice FIR/deconvolution results for the dataset-provided semantic similarity metric. The figure summarizes ROI-level significance and delayed lag profiles in original BOLD. The semantic similarity metric was significant in all 12 ROIs, whereas surprisal was not significant in any ROI after FDR correction.}
    \label{fig:alice_fir}
\end{figure}

In the Alice original-BOLD FIR/deconvolution analysis, the semantic similarity metric was significant in all 12 ROIs after FDR correction. In contrast, surprisal was not significant in any ROI after FDR correction. The semantic relevance effects were strong across the ROI set, with the smallest within-predictor FDR values observed in STG and MTG. The lag-level coefficients showed that semantic relevance effects were concentrated at delayed lags, consistent with a hemodynamic response profile rather than an instantaneous word-level artifact, as shown in Figure~\ref{fig:alice_fir}.

\subsubsection{Transformed-BOLD GAMMs Results}

\begin{figure}[t]
    \centering
    \includegraphics[width=\textwidth]{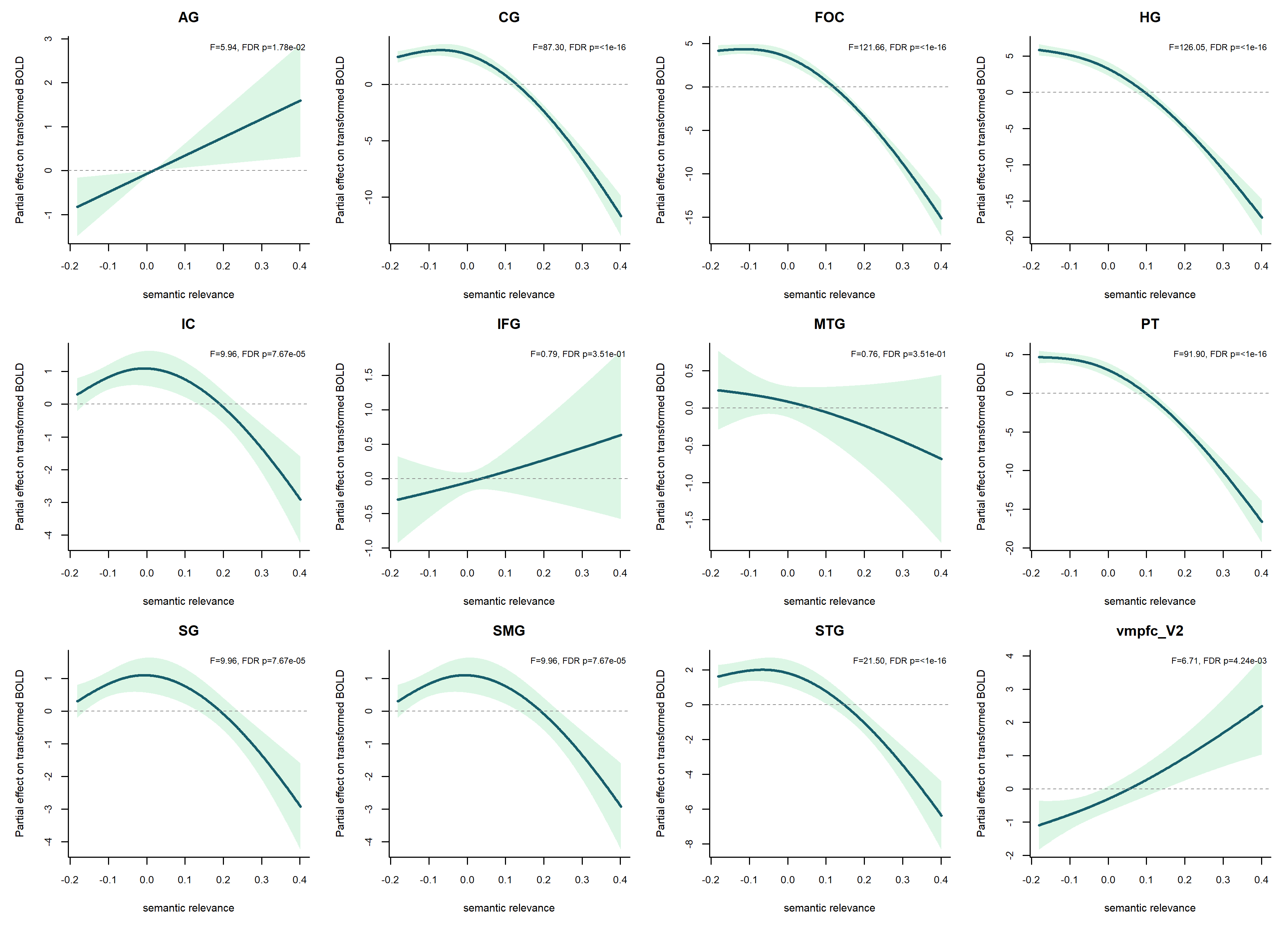}
    \caption{Partial effects of the dataset-provided semantic similarity metric on transformed BOLD in Alice. Each panel shows the GAMM-estimated smooth while controlling for log \texttt{wordfreq}, word length, duration, RMS intensity, prosodic break, surprisal, and participant effects.}
    \label{fig:alice_gamm}
\end{figure}

In the control-expanded Alice GAMMs, \texttt{semrel} remained FDR-significant in all 12 ROIs after log frequency, word length, duration, RMS intensity, prosodic break, surprisal, and participant were included. Surprisal was significant in 10/12 ROIs: AG, FOC, HG, IC, IFG, MTG, PT, SG, SMG, and STG; it did not survive correction in CG ($q=.132$) or vmPFC ($q=.347$). RMS was significant in 10/12 ROIs and prosodic break in 5/12. Thus, adding acoustic and prosodic controls did not attenuate the semantic-relevance result, but it also revealed broader conditional surprisal associations than the earlier model. The augmented model containing participant-specific random smooth yielded the same results for semantic relevance and surprisal respectively. Both predictors remained stable after allowing participant-level slope variation. The partial effect of semantic relevance in each ROI is illustrated in Figure~\ref{fig:alice_gamm}. Across the 12 ROIs, the estimated partial-effect curves for semantic relevance were heterogeneous rather than uniformly monotonic. The curves generally decreased across the observed semantic-relevance range in CG, FOC, HG, MTG, PT, and STG, whereas AG, IFG, and vmPFC showed increasing trends and IC, SG, and SMG exhibited  a different trend. The fitted curves indicated regionally heterogeneous linear and nonlinear associations rather than a single response shape shared across ROIs. %As the response was transformed and the model-level adjusted $R^2$ was extremely high, these GAMMs are interpreted as sensitivity and effect-shape analyses rather than independent predictive evidence.

In summary, in the Alice dataset, the dataset-provided semantic relevance metric showed robust associations with BOLD responses in both the transformed-BOLD GAMM analyses and the original-BOLD FIR/deconvolution analyses. In contrast, surprisal was not robustly associated with the original-BOLD responses in the FIR/deconvolution analyses, although it showed a significant association in the transformed-BOLD GAMMs. Overall, semantic relevance exhibited more consistent and stronger associations with BOLD activity than surprisal across analyses.

\subsection{Results of the Narratives Dataset}
\subsubsection{Full-window FIR/deconvolution omnibus analysis}

The Narratives FIR/deconvolution analysis included all 47 participants, 185 runs, and the continuous official clean residuals standardized within run. Semantic relevance, surprisal, log word frequency, word length, word duration, and word rate were represented by 11 FIR regressors spanning 0--15~s after word onset. Participant-specific story intercepts and linear and quadratic within-story time trends were residualized before pooling the TR-level observations. For each target predictor, the full model was compared with a reduced model from which all 11 FIR coefficients for that predictor had been removed. The resulting 11-df likelihood-ratio tests therefore evaluated whether the predictor exhibited an association with BOLD at any lag in the complete 0--15-s window. BH--FDR correction was applied across the 20 ROIs separately for semantic relevance and surprisal.

\begin{figure}[!h]
	\centering
	\includegraphics[width=\textwidth]{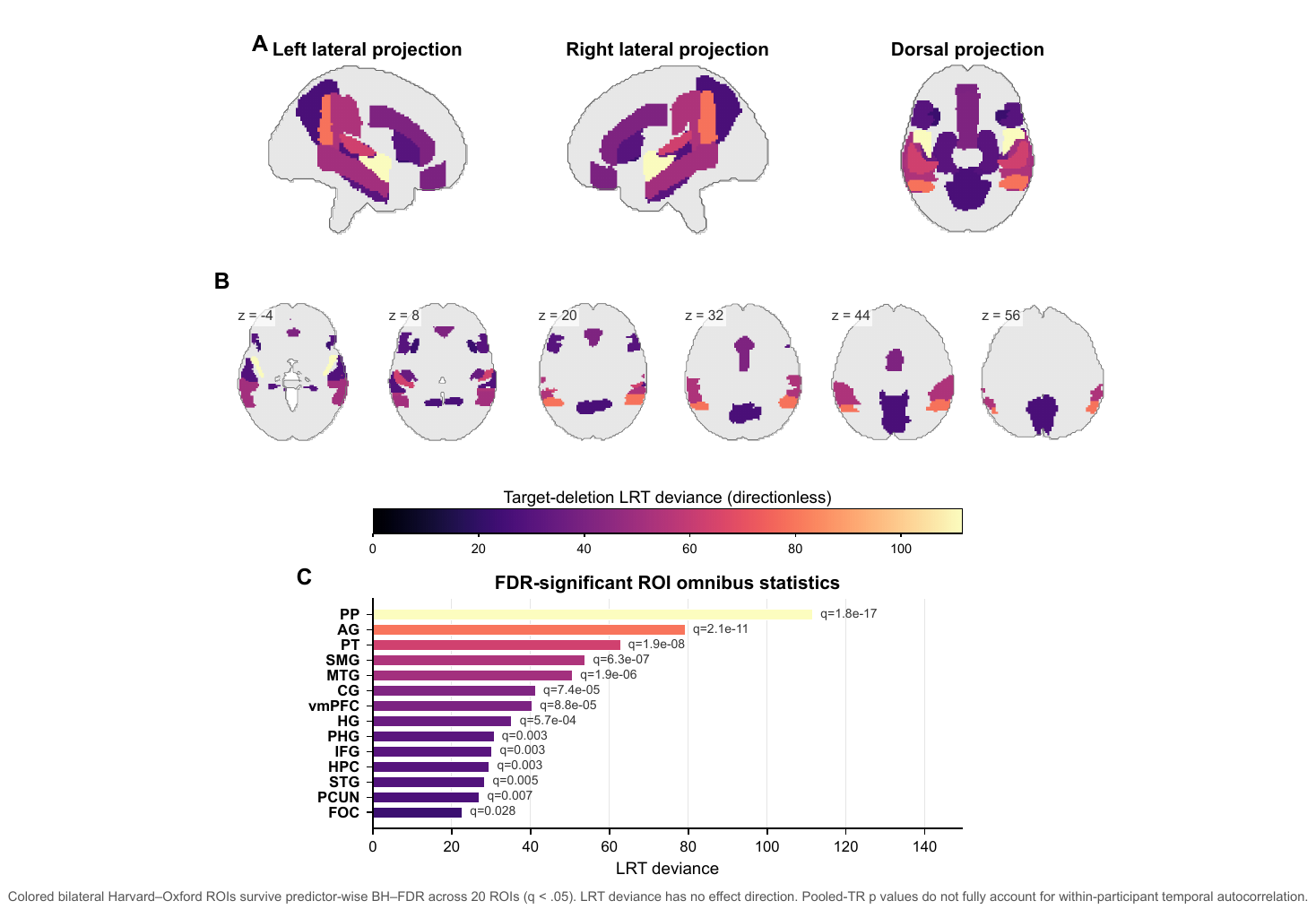}
	% \caption{Moth FIR/deconvolution brain-region results for semantic relevance. The plotted metric is \texttt{semantic relevance}. Warmer colors indicate stronger negative delayed effects in the HRF-weighted 4--12 s window. All 30 analyzable ROIs survived directional FDR correction.}
\caption{Narratives full-window FIR omnibus association for semantic relevance. The pooled TR-level target-deletion likelihood-ratio test jointly evaluated all 11 semantic-relevance FIR coefficients spanning 0--15~s in 47 participants and 185 runs. The full model additionally included surprisal, log word frequency, word length, word duration, and word rate, with participant-specific story intercepts and linear and quadratic time trends residualized. Of the 20 bilateral Harvard--Oxford ROIs tested, 14 survived predictor-wise BH--FDR correction ($q<.05$). (\textbf{A}) Left-lateral, right-lateral, and dorsal projections of significant ROIs. (\textbf{B}) Representative axial slices. (\textbf{C}) LRT deviance and FDR-adjusted $q$ value for each significant ROI. Colors encode the directionless target-deletion LRT deviance and therefore do not indicate the sign or temporal profile of the effect. Because this analysis pools TR-level observations, its $p$ values do not fully account for within-participant temporal autocorrelation.}
	\label{fig:narrative_fir}
\end{figure}

Semantic relevance survived BH--FDR correction in 14/20 ROIs: AG, CG, FOC, HG, HPC, IFG, MTG, PCUN, PHG, PP, PT, SMG, STG, and vmPFC. Surprisal survived correction in 15/20 ROIs: AG, FOC, FUS, HG, HPC, IFG, ITG, MTG, PCUN, PP, PT, SMG, STG, TP, and vmPFC. The minimum corrected values were $q=1.77\times10^{-17}$ for semantic relevance and $q=8.61\times10^{-224}$ for surprisal, as shown in Figure~\ref{fig:narrative_fir}. As a result, both predictors accounted for significant lagged BOLD variation across a broad portion of the language and narrative-comprehension network. Semantic relevance was nearly as spatially prevalent as surprisal, although the number of significant ROIs alone does not establish that one predictor provided a larger effect or greater unique predictive contribution.

Since the omnibus likelihood-ratio statistic tests the complete FIR coefficient vector, it does not provide an effect direction, and significant responses may contain positive coefficients, negative coefficients, or temporally varying combinations across the 0--15-s window. Accordingly, the resulting maps should be interpreted as maps of conditional omnibus association strength rather than positive or negative activation. In addition, because this analysis pooled TR-level observations, its likelihood-ratio $p$ values do not fully represent within-participant temporal autocorrelation. We therefore treat it as a full-window sensitivity analysis complementary to the more conservative participant-level analyses.

%\begin{table}[t]
%	\centering
%	\caption{Narratives full-window FIR/deconvolution omnibus results across 20 ROIs. Counts indicate ROIs surviving predictor-wise BH--FDR correction for the 11-df, 0--15-s target-deletion likelihood-ratio tests.}
%	\label{tab:narratives_fir_omnibus}
%	\begin{tabularx}{\textwidth}{lcX}
%		\toprule
%		Predictor & Significant ROIs & ROIs surviving FDR \\
%		\midrule
%		Semantic relevance
%		& 14/20
%		& AG, CG, FOC, HG, HPC, IFG, MTG, PCUN, PHG, PP, PT, SMG, STG, and vmPFC \\
		
%		Surprisal
%		& 15/20
%		& AG, FOC, FUS, HG, HPC, IFG, ITG, MTG, PCUN, PP, PT, SMG, STG, TP, and vmPFC \\
%		\bottomrule
%	\end{tabularx}
%\end{table}

\subsubsection{Transformed-BOLD GAMMs}

In the GAMM analyses, semantic relevance was significant after FDR correction in 17 of the 20 ROIs, whereas surprisal was significant in 16 of the 20 ROIs.
 Semantic relevance did not survive correction in FUS, HPC, or ITG; surprisal did not survive in AG, IC, or PHG. The parallel maximum-likelihood nested deletion tests produced valid FDR-significant evidence in 15/20 ROIs for semantic relevance and 14/20 for surprisal. Two semantic comparisons (SMG and dmPFC/SFG) and one surprisal comparison (IC) were invalid because the deletion comparison produced negative effective degrees of freedom or deviance changes, and these were not counted as null or significant findings. The partial effects of semantic relevance is illustrated in Figure~\ref{fig:narratives_sem_gamm}. The partial effect of surprisal is presented in Section 3 of the Appendix (Figure~\ref{fig:narratives_surp_gamm}).

\begin{figure*}[t]
    \centering
    \includegraphics[width=\textwidth]{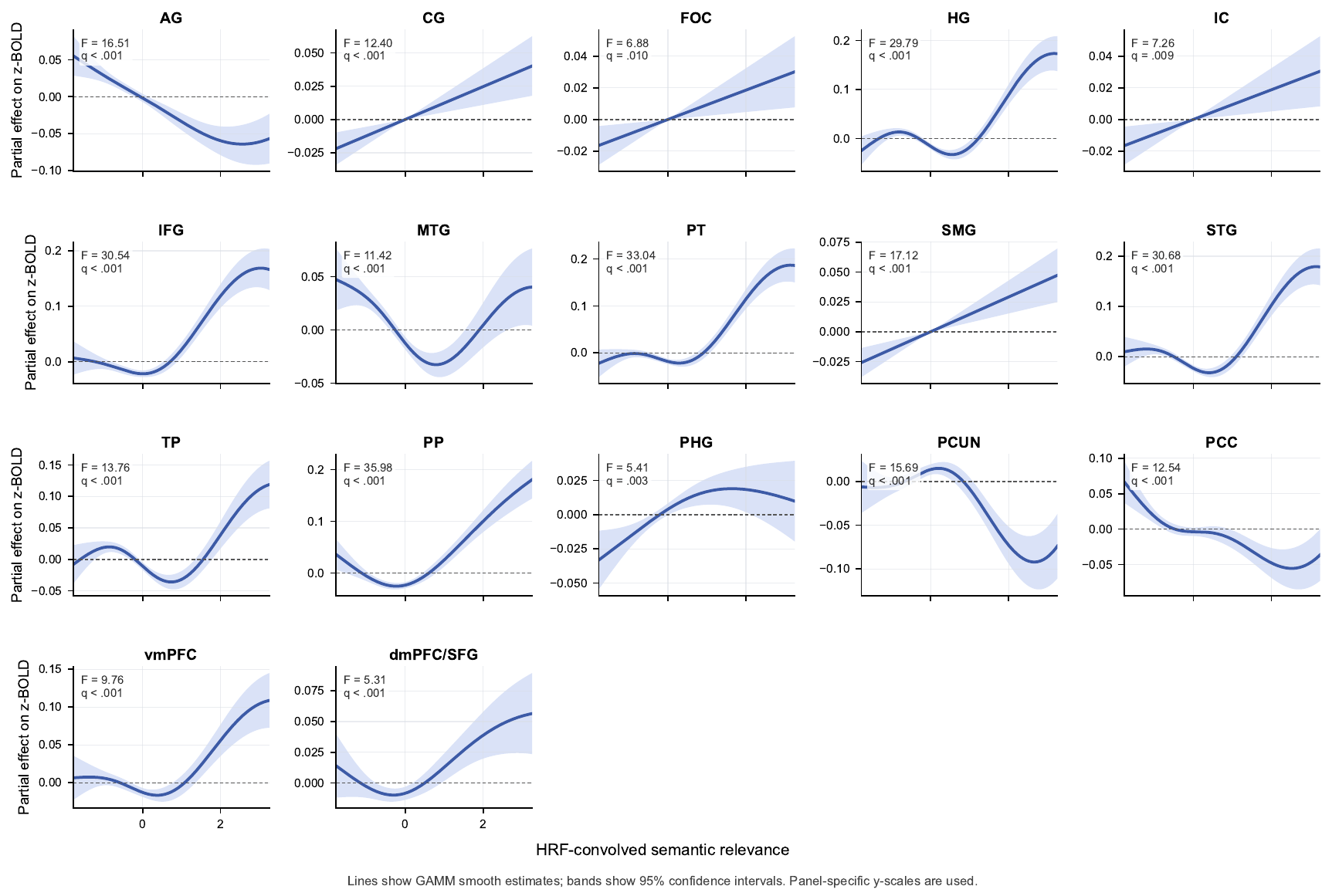}
    \caption{Narratives GAMM partial effects for semantic relevance in the 17 ROIs surviving BH--FDR correction in GAMM fittings. Curves show conditional smooth effects on run-standardized BOLD after controlling for HRF-convolved log frequency, word length, duration, word rate, surprisal, story-specific time, participant, and run. Panel labels report the smooth-term $F$ statistic.}
    \label{fig:narratives_sem_gamm}
\end{figure*}

Similar to Alice, the Narratives partial-effect curves likewise varied across ROIs in both direction and shape. Some ROIs showed increasing or decreasing monotonic associations, whereas others showed curved or non-monotonic profiles. Accordingly, an FDR-significant global smooth-term test indicated that the predictor was associated with transformed BOLD after adjustment for the other model terms, but did not imply a common direction or functional form across the Narratives ROI set. The partial effect of surprisal has the similar trends, as shown in Figure~\ref{fig:narratives_surp_gamm} of the Appendix.  Further, in Narratives, the augmented model containing participant-specific random smooth yielded FDR-significant semantic relevance and surprisal smooths in 17/20 and 17/20 ROIs, respectively. Both predictors remained stable
after allowing participant-level slope variation.

%The GAMM fittings did not explicitly model residual autocorrelation. In the stricter run-aware fREML plus AR(1) sensitivity analysis, the FDR-significant counts decreased to 8/20 for semantic relevance and 7/20 for surprisal. The reduction for both predictors shows that the 17/20 matched-model counts should be interpreted as conditional association and effect-shape evidence, not as independent or cross-validated prediction. Across both GAMM specifications, there was no basis for claiming that semantic relevance was more spatially prevalent than surprisal.

%\subsection{Cross-dataset comparison}

The two statistical methods yielded a qualified cross-dataset pattern. Alice provided the clearest semantic-relevance result in the timing-sensitive analysis because 12/12 semantic-relevance FIR effects and 0/12 surprisal FIR effects. Its expanded GAMMs, however, showed that surprisal was conditionally associated with transformed BOLD in 10/12 ROIs once RMS intensity and prosodic break were controlled. Narratives did not reproduce a broad semantic advantage. The GAMMs identified 17/20 and 16/20 ROIs for each predictor. Accordingly, the data support BOLD sensitivity to both contextual semantic fit and probabilistic unexpectedness, but not a general ranking in which semantic relevance dominates surprisal across datasets and statistical methods.

\subsection{Additional Results}

First, we summarized the GAMM results using the smooth-term $F$ statistics as descriptive measures of conditional association strength. In the updated Alice transformed-BOLD GAMMs, semantic relevance survived BH--FDR correction in all 12 ROIs (median $F=9.81$, range: 3.85--30.83), whereas surprisal survived in 10/12 ROIs (median $F=5.18$, range: 1.51--15.28). In the Narratives HRF-convolved GAMMs, both semantic relevance and surprisal had FDR-significant smooth terms in 17/20 ROIs, with median $F$ statistics of 12.47 (range: 2.07--35.98) and 13.65 (range: 0.07--277.71), respectively. Complementary nested model comparisons identified significant unique contributions in 15/20 ROIs for semantic relevance and 14/20 ROIs for surprisal. Model-level fit was substantially higher in Alice (median deviance explained $\approx .999$) than in Narratives (median deviance explained $=.0055$; median adjusted $R^2=.0052$).

Second, across the four stories in Narratives, mean lag-1 autocorrelation was $-.148$ for semantic relevance and $.055$ for surprisal, with both predictors approaching zero over subsequent word lags. These results indicate distinct short-range temporal structures for the two predictors but do not support characterizing semantic relevance as a uniformly slower or more positively autocorrelated predictor (Figure~\ref{fig:smoothness}, Appendix~\ref{otherresults}).

Third, FIR/deconvolution analyses of 11 anatomically corresponding ROIs compared temporal response profiles between Alice and Narratives. In Alice, semantic relevance survived the grouped full-window test in all common ROIs, whereas surprisal did not survive FDR correction. In Narratives, semantic relevance and surprisal were significant in 10/11 and 9/11 common ROIs, respectively. Thus, both predictors showed distributed associations with BOLD responses in Narratives, although the temporal response profiles differed from those observed in Alice (Figure~\ref{fig:common-roi-fir}, Appendix~\ref{otherresults}).

Finally, we conducted a supplementary descriptive functional-connectivity analysis of the Narratives dataset. The strongest correlation was observed between HG and PT ($r=.805$), with additional prominent connections linking auditory-language, semantic, episodic-memory, and default-mode regions. These connectivity analyses were descriptive only, based on a visualization threshold, and were not used for statistical inference regarding semantic relevance or surprisal (Appendix~\ref{connectivity}).

\section{Discussion}

Our results show that, in the Alice dataset, semantic relevance was robust in both the FIR/deconvolution and GAMM analyses, whereas surprisal was detected in the GAMMs. In the Narratives dataset, both semantic relevance and surprisal showed reliable associations with BOLD responses. Overall, both contextual semantic relevance and surprisal were associated with BOLD activity, although their apparent strength depended on the dataset and analytical approach. Semantic relevance nevertheless exhibited the more consistent pattern of associations across datasets in the timing-sensitive FIR analysis: it was significant in all 12 Alice ROIs (versus 0 for surprisal) and in 17/20 Narratives ROIs (versus 16/20 for surprisal). In the GAMM analyses, the two predictors were comparably prevalent in both datasets. The implications of these findings are discussed below.

\subsection{Semantic relevance as sustained semantic integration}

Semantic relevance differs from surprisal in both computational meaning and expected neural timescale. Surprisal measures how unexpected a word is given its preceding context. Semantic relevance instead measures how strongly the current word fits with, relates to, or contributes to the recent semantic context. Thus, a word can be highly surprising but still semantically appropriate, and a predictable word may contribute little to the developing discourse representation.

This distinction helps explain why semantic relevance may be robust in BOLD. The BOLD response is slow, delayed, and temporally smoothed \citep{boynton1996linear,logothetis2003bold}. It may be therefore less suited to isolating brief word-by-word prediction errors, but well suited to detecting processes that unfold over several seconds. The temporally overlapping local-context estimates may track contextual semantic structure that changes continuously during comprehension. This kind of sustained contextual processing is compatible with previous naturalistic fMRI work showing that semantic information is represented across distributed cortical systems during story comprehension \citep{huth2016semanticmaps,wehbe2014harrypotter,jain2018context}.

The same account also links the present fMRI findings to EEG and eye-movement evidence. In EEG, semantic fit and contextual integration are often reflected in N400-family responses, which are sensitive to how easily a word can be integrated with prior context \citep{kutas1984expectancy,kutas2011n400,frank2015erp}. In eye movements and reading-time measures, semantically well-fitting words are expected to reduce processing difficulty, whereas weakly fitting words may increase fixation time, regressions, or later integration cost. These measures capture faster behavioral and EEG consequences of semantic fit, and BOLD captures the slower hemodynamic consequence of related integration and updating processes. In this sense, semantic relevance can plausibly affect EEG, eye movements, and BOLD, but with different temporal signatures.

The sign of the effect also remains an empirical question. In Narratives FIR/deconvolution, semantic relevance was negative in AG and MTG but positive in CG, IFG, PP, and PT. Negative associations can be consistent with facilitated integration or reduced updating demand \citep{grill-spector2006repetition}, whereas positive associations may reflect stronger semantic binding, retrieval, or contextual updating. The mixed directions argue against assigning a universal ``integration effort'' interpretation to the metric and show why a two-sided participant-level test is preferable to a post hoc directional test.

\subsection{Neural mechanisms underlying semantic relevance effects}

The regionally heterogeneous direction of semantic-relevance effects observed in the present data can be interpreted through the Memory, Unification, and Control (MUC) framework, which proposes that language comprehension recruits distinct but interacting neural operations \citep{hagoort2013muc}. In this framework, negative semantic-relevance associations in AG and MTG may reflect reduced retrieval demand or neural adaptation when the incoming word is semantically congruent with the recent context: a well-fitting word requires less effortful access to stored semantic representations \citep{grill-spector2006repetition,binder2011semanticmemory,seghier2013angular}. Conversely, positive associations in IFG and FOC are consistent with unification operations, in which a semantically relevant word triggers stronger combinatorial processing as it is actively bound into the evolving syntactic and semantic structure \citep{hagoort2013muc}. The positive effects in CG may reflect increased cognitive control or conflict monitoring accompanying deeper contextual integration. Thus, the same metric, semantic relevance, can produce opposing BOLD directions across regions because it engages functionally distinct operations: facilitated retrieval in posterior semantic-storage regions and enhanced binding in frontal unification regions. This interpretation remains tentative, as ROI-level BOLD associations cannot isolate specific neural computations, but it provides a mechanistically grounded account of why a single predictor should not be expected to produce uniform activation across the language network.

The distinction between surprisal and semantic relevance also maps onto a hierarchical organization of cortical information processing. According to the predictive coding framework, the brain maintains generative models at multiple hierarchical levels, with prediction errors propagated bottom-up and contextual predictions sent top-down \citep{rao1999predictive,friston2005theory}. Surprisal, as a local probabilistic prediction error, may primarily drive bottom-up error signals that are rapid, transient, and strongest in early processing stages. Semantic relevance, in contrast, reflects the degree of alignment between incoming input and a maintained contextual representation, a computation more consistent with top-down contextual modulation. This directional distinction is supported by the present finding that semantic relevance showed its clearest advantage over surprisal in the timing-sensitive Alice FIR analysis, where its effects followed a delayed hemodynamic profile consistent with sustained processing rather than rapid error signaling. The hierarchical temporal receptive window framework further predicts that higher-order regions, such as AG, PCUN, PCC, and vmPFC, integrate information over longer timescales during naturalistic comprehension \citep{hasson2008hierarchy,lerner2011temporal,hasson2015hierarchical}. The fact that semantic relevance was significant in these regions across both datasets is compatible with the view that it captures a contextual integration signal that is expressed preferentially at the temporal scales and cortical levels where top-down narrative-level representations are maintained and updated \citep{simony2016dynamic,baldassano2017discovering}.

Finally, the computational form of semantic relevance, a distance-weighted combination of cosine similarities between distributed word vectors, has a natural correspondence with neural population-level similarity computations. Representational similarity analysis has shown that the geometric relationships among neural activity patterns in language regions mirror the semantic structure of the stimuli \citep{kriegeskorte2008representational,huth2016semanticmaps}. Semantic relevance can be understood as measuring the alignment between the population-level representation of the incoming word and the representations of recently processed words, weighted by recency. This formulation is consistent with evidence that cortical language regions maintain distributed semantic representations that are continuously updated during naturalistic comprehension \citep{wehbe2014harrypotter,jain2018context,caucheteux2022brains}. The association between semantic relevance and BOLD in regions spanning temporal, parietal, frontal, and medial systems, including HPC and PHG, further suggests that semantic-fit computations may interface with episodic memory binding mechanisms \citep{ranganath2012twocortical}. During story listening, the hippocampal system is thought to bind incoming semantic content with the ongoing situation model, and the present finding that HPC and PHG showed significant semantic-relevance effects in Narratives is compatible with this account. However, the present analyses cannot determine whether the hippocampal associations reflect direct semantic-fit computations or downstream consequences of cortical integration, and this distinction remains an important target for future connectivity-based or laminar-resolution investigations.

\subsection{Surprisal effects are dataset- and model-dependent}

Surprisal is still taken as a central metric of local probabilistic prediction and processing difficulty \citep{hale2001earley,levy2008surprisal}. Its absence from the Alice FIR test should not be generalized to fMRI as a modality because surprisal survived participant-level FDR correction in 15/20 Narratives ROIs and had broad conditional effects in GAMMs in Narratives. These findings are consistent with prior reports of surprisal-related BOLD in temporal and inferior frontal regions \citep{willems2016prediction,brennan2016abstract,shain2020fmri}.

Several factors can nevertheless make surprisal estimates dataset-dependent. The predictor changes with the language model, tokenizer, and available context; its unique variance depends on lexical, timing, acoustic, and semantic covariates; and adjacent word responses overlap in BOLD. The Alice result also shows that adding controls need not make surprisal less significant. After RMS and prosodic break were added, the independent smooth term was detectable in more ROIs. This pattern is compatible with suppression or redistribution of shared variance and cautions against interpreting the number of significant ROIs as a direct measure of computational importance.

Studies using different stimuli, HRF assumptions, control sets, and analysis levels may therefore obtain different surprisal results. In the current data, surprisal is not generally weaker than semantic relevance, and it is robust in Narratives and selectively absent from one timing-sensitive Alice analysis.

%\subsection{Why Alice and Narratives differ}

\subsection{Prediction and semantic integration as complementary computations}

%Alice comprises a single, identically presented literary narrative, whereas the analyzed Narratives sample includes four stories, 185 runs, and substantially greater between-story and between-run variability \citep{bhattasali2020alice,nastase2021narratives}. Accordingly, Alice provides tighter temporal and discourse alignment, whereas Narratives offers a larger, more heterogeneous sample that requires explicit modeling of story, run, and temporal structure. The datasets also differed in their semantic-relevance measures and response representations. Alice used the dataset-provided semantic relevance and transformed BOLD response, whereas Narratives used a fixed Model2Vec-based semantic relevance metric and nuisance-cleaned, within-run standardized BOLD signals. Thus, the observed differences likely reflect methodological differences between datasets rather than intrinsic differences in semantic processing, highlighting the need for future analyses using a common semantic-relevance metric across both datasets.

The present findings speak to a central debate in language comprehension: whether neural responses during comprehension primarily reflect prediction, integration, or the interaction between the two. Surprisal provides an information-theoretic measure of local probabilistic expectation, indexing how unexpected a word is given the preceding context \citep{hale2001earley,levy2008surprisal}. Prediction-based accounts propose that comprehenders use context to pre-activate likely upcoming linguistic input, but the extent and level of prediction remain debated \citep{kuperberg2016prediction,pickering2018predicting}. Semantic relevance, in contrast, indexes how well the incoming word fits the recent semantic context. It is therefore closer to semantic integration or contextual fit than to local probability error.

This distinction helps clarify why surprisal and semantic relevance should not be treated as competing versions of the same predictor. A word can be locally unexpected but still semantically coherent, and a predictable word can add little to the current semantic representation. Surprisal may therefore capture the cost of local expectation violation or lexical access, whereas semantic relevance may capture how easily a word can be incorporated into an evolving contextual representation. This view is consistent with models in which language comprehension depends on multiple interacting operations, including memory retrieval, unification, prediction, and control \citep{hagoort2013muc,lau2008n400}.

The present results support a complementary-computation interpretation. Semantic relevance does not replace surprisal, and neither predictor consistently dominates the other. Instead, each retained conditional associations when the other was included, consistent with partly distinct contributions from contextual semantic fit and probabilistic expectation. %Because no held-out prediction comparison was performed here, these associations do not yet establish unique out-of-sample predictive gain.
An open question is whether prediction and integration interact at the level of individual words. Words that are both highly surprising and semantically relevant, that is, unexpected yet contextually fitting, may elicit a distinctive neural response reflecting simultaneous prediction error signaling and successful integration. Conversely, words that are predictable but semantically peripheral may produce low surprisal without engaging deep contextual binding. The present analyses tested the conditional main effects of each predictor but did not model their interaction. Future work could examine whether surprisal $\times$ semantic relevance interaction terms explain additional BOLD variance, particularly in regions such as IFG and AG where both predictors showed significant effects. Such analyses would help determine whether the two computations operate independently or whether their neural expression depends on the joint informational profile of the incoming word.

\subsection{Fast EEG responses and slow hemodynamic timescales}

An implication concerns the temporal scales at which lexical expectation and contextual semantic fit become detectable. Surprisal and predictability effects are well established in temporally precise measures such as EEG and eye movements. In EEG, predictable or semantically fitting words typically elicit reduced N400 amplitudes, linking contextual expectation and semantic access to neural responses within a few hundred milliseconds after word onset \citep{kutas1984expectancy,kutas2011n400,lau2008n400,frank2015erp}. Eye-movement studies similarly show that word predictability and contextual constraint influence fixation durations and other reading behaviors during incremental comprehension \citep{rayner1996contextual,smith2013surprisal, sun2026attention}.

A companion naturalistic-reading EEG analysis provides converging evidence \citep{sun2026eeg}. Across 22 participants and 32 channels, both semantic relevance and surprisal were associated with N400 and P600 responses, but with partly distinguishable temporal profiles: semantic relevance showed broader effects in the N400 window, whereas surprisal exhibited comparatively stronger modulation around the P600 interval. The weak correlation between the two metrics in that dataset further suggests that they capture partly distinct aspects of word-level processing.

These EEG results indicate that semantic relevance should not be understood exclusively as a slow process. Contextual semantic fit can influence neural activity within several hundred milliseconds, affecting both early semantic access or integration and later updating or reanalysis. At the same time, EEG preserves rapid word-locked fluctuations that may be attenuated in fMRI. BOLD responses are delayed and temporally smoothed \citep{boynton1996linear,logothetis2003bold}, and in continuous speech the hemodynamic responses elicited by adjacent words overlap substantially. A brief and rapidly varying surprisal-related response may therefore be difficult to isolate in BOLD, particularly after controlling for correlated lexical variables such as frequency, word length, and duration.

Semantic relevance may be comparatively well aligned with the temporal characteristics of BOLD because it is computed from overlapping relations between each target word and its recent context. Although each estimate is word-specific, adjacent estimates share contextual information and may track semantic structure that evolves continuously over several words. Such temporally extended contextual variation may survive hemodynamic smoothing more readily than highly local fluctuations in probabilistic unexpectedness. This interpretation is compatible with evidence for hierarchical temporal receptive windows across cortex, whereby early auditory and language regions respond to relatively short-timescale information, whereas higher-order temporal, parietal, and default-mode regions integrate information over longer periods during naturalistic comprehension \citep{hasson2008hierarchy,lerner2011temporal,hasson2015hierarchical}.

In sum, the EEG and fMRI findings suggest differences in measurement sensitivity rather than a strict separation between fast and slow linguistic computations. Both surprisal and semantic relevance can modulate rapid electrophysiological responses, and the Narratives results show that both can also be recovered from slow BOLD. Their estimated prevalence remains dependent on stimulus structure, predictor construction, temporal alignment, controls, and residual modeling.

\subsection{Naturalistic semantic networks and situation-model updating}

The use of naturalistic spoken narratives is central to the present study. Real story comprehension requires listeners to maintain and update characters, events, goals, causal relations, and discourse context over time. Naturalistic fMRI studies have shown that semantic information is represented across distributed cortical systems rather than being confined to classical language areas \citep{huth2016semanticmaps,wehbe2014harrypotter,jain2018context}. This broader organization makes naturalistic datasets especially useful for testing predictors related to contextual semantic fit.

Semantic relevance was associated with activity across a distributed comprehension network, although its spatial extent depended on the inferential test. In the pooled 0--15-s FIR omnibus analysis, jointly removing all 11 semantic-relevance lag coefficients significantly reduced model fit in 14/20 ROIs. This broader result indicates that semantic relevance was associated with temporally distributed BOLD variation beyond the restricted 4.5--12-s contrast. %However, because the omnibus statistic is directionless, it does not establish whether an ROI showed a positive or negative response or identify the lag at which the association arose.

The anatomical distribution of these effects is compatible with the involvement of processes supporting lexical-semantic access, semantic selection, contextual integration, episodic or discourse-level memory, and situation-model updating \citep{ferstl2008neuroanatomy}. Nevertheless, this interpretation remains functional rather than process-specific. Anatomical ROI averages cannot determine which computation generated an association, and semantic relevance may covary with unmodeled aspects of narrative structure, attention, memory demand, or discourse progression.
%The GAMM effects should be interpreted more cautiously because their statistical extent was sensitive to AR(1) modeling. Likewise, the pooled FIR omnibus LRT treated TR-level observations as pooled and therefore did not fully represent within-participant temporal autocorrelation. Consequently, the participant-level FIR analysis provides the stronger inferential basis for claims about reliable delayed effects across listeners, whereas the pooled omnibus and GAMM results provide complementary evidence about the possible spatial extent and temporal distribution of semantic-relevance associations. 
In short, the results support a distributed but regionally heterogeneous relationship between semantic relevance and narrative-evoked BOLD responses, rather than a uniform network-wide semantic-relevance response.

\subsection{Methodological implications and limitations}
\label{sec:limitations}

The results have broader implications for the computational neuroscience of language. %Computational predictors should not be treated as interchangeable regressors simply because they are derived from the same words. Surprisal quantifies probabilistic unexpectedness under a language model, whereas semantic relevance operationalizes the degree of semantic fit between the current word and its preceding context. The two metrics were only weakly correlated in both datasets and therefore captured partly distinct properties of the linguistic input. Their neural associations may also unfold over different timescales, making the choice of temporal model important.
Recent work has demonstrated that GPT-based language models can predict sentence-level responses in the human language network and can identify novel sentences that reliably drive or suppress its activity, with surprisal and linguistic well-formedness contributing to response strength \citep{Tuckute2024network}. These findings illustrate the value of computational representations for predicting neural responses to language. Nevertheless, successful prediction by a computational representation does not establish that the brain implements the model's training objective or computational mechanism \citep{Antonello2024predict}. Accordingly, the present findings should be interpreted as evidence that BOLD responses are sensitive to information captured by semantic relevance, rather than as proof that the brain explicitly computes the metric in precisely the form implemented here. Moreover, although semantic relevance showed particularly widespread effects in Alice, its relative advantage over surprisal was analysis-dependent in Narratives. The results therefore support partially distinct neural associations rather than an unconditional superiority of one predictor.

The complementary use of GAMMs and FIR/deconvolution helps separate different aspects of these associations. GAMMs applied to transformed or HRF-convolved predictors provide flexible tests of predictor--response relationships after controlling measured lexical, acoustic, rate, and participant-level variation. Although the smooth terms permit nonlinear relationships, the estimated effects need not be strongly nonlinear. FIR/deconvolution applied to the original continuous BOLD series instead represents predictor effects across multiple post-onset lags without imposing a single canonical HRF shape. The pooled full-window FIR test assesses whether the complete lag vector improves model fit across the pooled TR-level data.

This distinction is important because BOLD is intrinsically low-pass and semantic relevance performs differently from surprisal. A smoother predictor may align more readily with slow BOLD fluctuations even in the absence of a uniquely semantic mechanism. The delayed FIR profiles, comparison with surprisal, and inclusion of lexical and timing controls reduce this concern but do not eliminate it. Temporal-spectrum-matched null models or other autocorrelation-preserving tests would provide a more stringent assessment of whether the semantic-relevance effects exceed those expected from predictor smoothness alone.

Recent work has introduced methods for estimating semantic relevance and has reported effects on eye movements during reading across multiple languages \citep{sun2024attention1,sun2026attention}. Related studies suggest that semantic relevance may also contribute to models of speech production, spontaneous speech, EEG responses, and visual attention \citep{sun2026speech,sun2026eeg,sun2024visual}. The present study extends this line of research by showing that semantic relevance is associated with BOLD responses during naturalistic language comprehension in two independent datasets. Whether the construct generalizes to sequential processing in other domains, such as decision-making, emotional dynamics, or action selection, remains a question for direct empirical testing.

Several limitations should be noted. First, the analyses are observational and cannot establish causal relationships between semantic relevance and BOLD responses. Second, the Narratives results were sensitive to temporal-dependence modeling, and the pooled FIR analysis did not fully account for within-participant autocorrelation. Third, ROI-based analyses may obscure finer-grained hemispheric, regional, and voxel-level effects.

\section{Conclusion}

%Across Alice and Narratives, semantic relevance and surprisal both showed conditional associations with BOLD during naturalistic speech comprehension. The results therefore favor complementary, dataset-dependent neural correlates of contextual semantic fit and probabilistic expectation over a claim that one metric generally dominates the other. Naturalistic fMRI remains valuable for testing language computations over continuous discourse. Until then, semantic relevance should be regarded as a promising complementary predictor of contextual processing, not as a demonstrated replacement for surprisal.
Across two naturalistic fMRI datasets, both semantic relevance and surprisal were associated with BOLD responses during spoken language comprehension. Semantic relevance showed the more consistent pattern across datasets and analysis methods, particularly in the Alice FIR analysis, but in Narratives the two predictors showed comparable spatial prevalence and effect strength. These findings suggest that contextual semantic fit and local probabilistic expectation make partially distinct, dataset-dependent contributions to hemodynamic responses during naturalistic listening.
%Across two naturalistic fMRI datasets, semantic relevance measures were associated with BOLD responses during spoken language comprehension, whereas surprisal showed weaker and less consistent effects in these ROI-level analyses. These findings suggest that contextual semantic fit may be readily expressed in slow hemodynamic responses. 
%The result is theoretically important because it links computational metrics of language comprehension to neural measurement timescales. Surprisal may primarily capture fast prediction-related processing, whereas semantic relevance may capture more sustained contextual integration. More broadly, the study highlights the value of naturalistic fMRI for examining language processes that unfold over continuous discourse. Semantic relevance may provide a useful bridge between word-level computational models and distributed neural systems supporting semantic integration, narrative comprehension, and situation-model updating. %At the same time, the findings should be treated as motivating evidence rather than a final confirmatory test. Future work should use a fixed independently selected semantic relevance metric, precise word-level timing, richer nuisance controls, temporal surrogate analyses, and cross-validated encoding models to determine whether semantic relevance explains delayed BOLD dynamics beyond lexical surprisal.
The results are theoretically important because they link computational metrics of language comprehension to the timescale of neural responses. Surprisal may primarily capture rapid prediction related processing, whereas semantic relevance may capture more sustained contextual integration. More broadly, this study highlights the value of naturalistic fMRI for investigating language processes that unfold during continuous discourse. Semantic relevance may provide a useful bridge between computational models operating at the word level and distributed neural systems supporting semantic integration, narrative comprehension, and situation model updating. Several directions could strengthen this account, including voxelwise encoding models that quantify the unique predictive contribution of semantic relevance beyond surprisal and lexical controls in held out data, contextualized embeddings to test whether context dependent representations improve BOLD prediction relative to the static embeddings used here, cross linguistic replication to examine whether the temporal and semantic structure of different languages modulates the relative sensitivity of the two metrics, and combined EEG and fMRI or MEG analyses to simultaneously resolve the rapid electrophysiological and slower hemodynamic signatures of language processing.

\section*{Data and Code Availability}

All analyses used publicly available fMRI datasets. Alice was obtained from the Alice dataset release \citep{bhattasali2020alice}; Narratives was obtained from OpenNeuro dataset ds002345 and its official derivatives \citep{nastase2021narratives}. Analysis scripts, result tables, and generated figures will be made available in a public repository upon publication. % TODO: Replace with actual repository URL before submission.

\section*{Acknowledgments}

We thank the creators of the Alice and Narratives datasets for making their naturalistic fMRI resources publicly available. Funding and contributor statements should be added before submission.

\bibliographystyle{apalike}
\bibliography{reference}

\pagebreak
\appendix

\section{ROI Definitions}
\label{app:rois}

\begin{table*}[htbp]
\centering
\caption{Brain regions included in the Alice and Narratives ROI analyses and their commonly associated functions. The functional descriptions are intentionally broad because each region contributes to multiple cognitive processes.}
\label{tab:roi_functions}
\small
\begin{tabularx}{\textwidth}{llX}
\toprule
\textbf{ROI} & \textbf{Full name} & \textbf{Commonly associated functions} \\
\midrule

\multicolumn{3}{l}{\textit{ROIs included in both Alice and Narratives}} \\
\addlinespace

AG & Angular gyrus &
Semantic and conceptual integration, discourse comprehension, and multimodal information processing. \\

CG & Cingulate gyrus &
Attention, cognitive control, conflict monitoring, and internally directed processing. \\

FOC & Frontal opercular cortex &
Speech production, articulation, phonological processing, and language-related cognitive control. \\

HG & Heschl's gyrus &
Primary auditory processing and early analysis of acoustic information. \\

IC & Insular cortex &
Salience detection, auditory--motor integration, speech processing, and cognitive control. \\

IFG & Inferior frontal gyrus &
Syntactic processing, semantic selection, language control, and speech production. \\

MTG & Middle temporal gyrus &
Lexical--semantic processing, conceptual representation, and sentence comprehension. \\

PT & Planum temporale &
Higher-order auditory analysis, phonological processing, and speech perception. \\

SMG & Supramarginal gyrus &
Phonological processing, verbal working memory, and multimodal integration. \\

STG & Superior temporal gyrus &
Speech perception, auditory-language comprehension, and semantic processing. \\

vmPFC & Ventromedial prefrontal cortex &
Schema-based interpretation, valuation, contextual integration, and internally guided cognition. \\

\midrule
\multicolumn{3}{l}{\textit{Additional ROI included only in Alice}} \\
\addlinespace

SG & Superior gyrus &
Higher-order sensory, auditory, or language-related processing, depending on the anatomical definition. \\

\midrule
\multicolumn{3}{l}{\textit{Additional ROIs included only in Narratives}} \\
\addlinespace

TP & Temporal pole &
Semantic, social-conceptual, emotional, and narrative-level integration. \\

ITG & Inferior temporal gyrus &
Object recognition, lexical-semantic representation, and conceptual processing. \\

FUS & Fusiform gyrus &
Visual-form processing, object recognition, and higher-level conceptual representation. \\

PP & Posterior parietal region &
Attention, working memory, and multimodal integration. \\

PHG & Parahippocampal gyrus &
Contextual memory, scene processing, and associations between events and environments. \\

HPC & Hippocampus &
Episodic memory, contextual association, and formation of event representations. \\

PCUN & Precuneus &
Episodic imagery, internally directed cognition, and construction of situation models. \\

PCC & Posterior cingulate cortex &
Internally directed cognition, episodic memory, and narrative situation-model processing. \\

dmPFC/SFG & Dorsomedial prefrontal / superior frontal cortex &
Social and narrative inference, executive control, working memory, and internally directed cognition. \\

\bottomrule
\end{tabularx}

\begin{minipage}{0.98\textwidth}
\footnotesize
\textit{Note.} Functional descriptions summarize commonly reported associations and should not be interpreted as exclusive functions. Narratives ROIs were bilateral Harvard--Oxford atlas composites (maximum-probability threshold 25\%, 2-mm template).
\end{minipage}
\end{table*}

\begin{figure}[!h]
	\centering
	\includegraphics[width=0.86\textwidth]{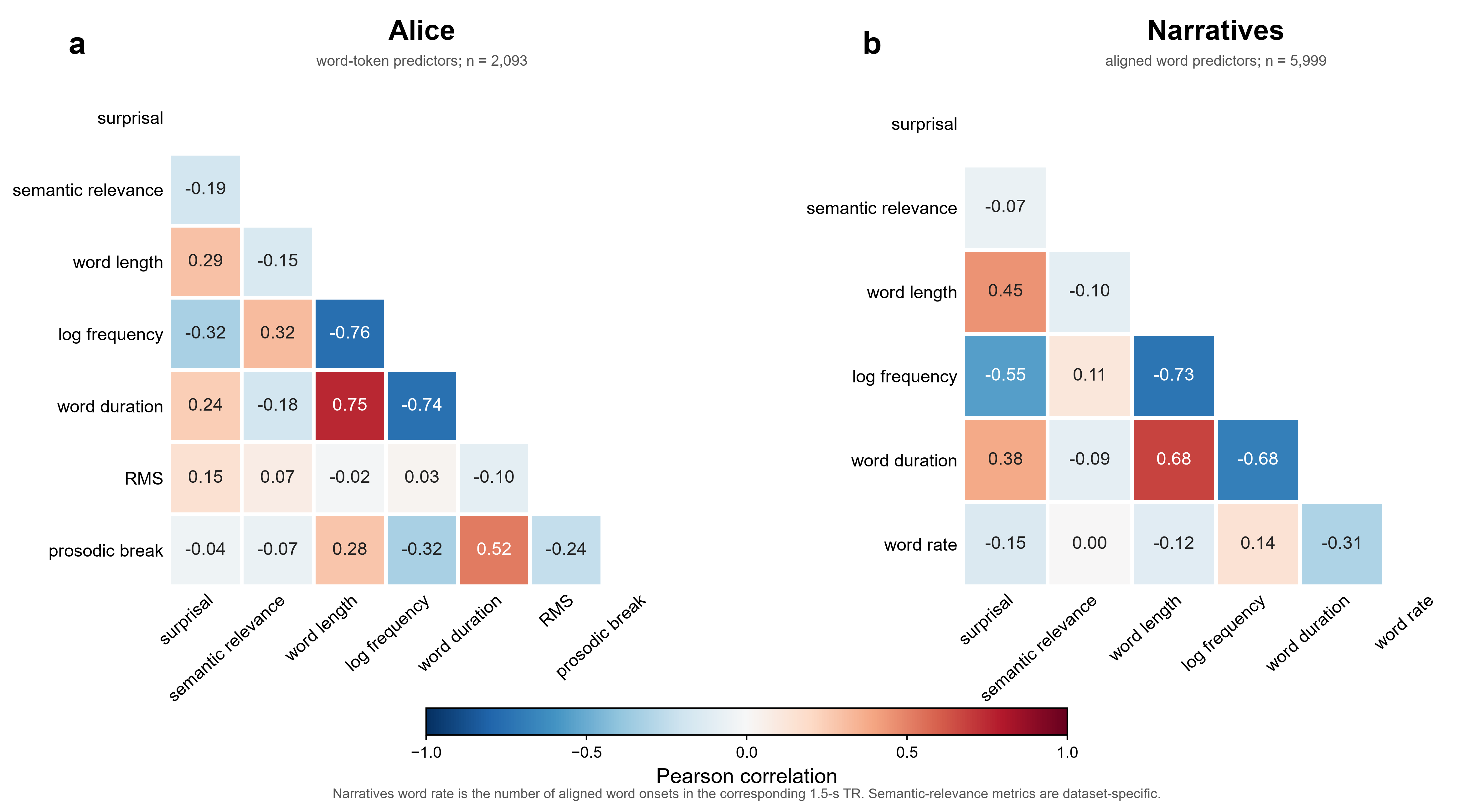}
	\caption{Correlations for predictors in the two datasets}
	\label{fig:corr}
\end{figure}

As shown in Figure~\ref{fig:corr}, surprisal is weakly correlated with semantic relevance in both datasets. Specifically, $r = -0.19$ in Alice, $r = -0.07$ in Narratives, and these indicate that surprisal and semantic relevance are distinct metrics used in the present study. 

\section{Robustness Analysis: Temporal Structure of Semantic Relevance}
\label{app:semrel-smoothness}

One potential concern is that a computational predictor may appear to explain BOLD partly because its temporal structure is well matched to the slow, low-pass characteristics of the hemodynamic response. This concern is particularly relevant when comparing semantic relevance with surprisal: if semantic relevance contains more slowly varying information, an apparently stronger BOLD association might reflect temporal-spectrum matching rather than semantic processing.

We therefore quantified the temporal structure of the predictors in both datasets. For Alice, the diagnostics were computed over the word-level predictor sequence from the single story. For Narratives, diagnostics were first computed separately for each of the four stories at the word level and then averaged across stories. As the Narratives GAMMs used predictors placed at their true word onsets, convolved with a canonical HRF, and sampled at TR resolution, we additionally calculated the same diagnostics for these HRF-convolved TR-level predictors. This distinction is important because HRF convolution substantially changes the temporal spectrum of both predictors.

\subsection{Temporal autocorrelation and spectral properties}

For each predictor series, we calculated the lag-1 autocorrelation, an AR(1)-based effective-sample-size diagnostic, the proportion of spectral power contained in the lowest 10\% of the positive-frequency range, and the spectral centroid. The effective-sample-size diagnostic was calculated as

\begin{equation}
	N_{\mathrm{eff}} = N \frac{1-r_1}{1+r_1},
\end{equation}

where $r_1$ is the lag-1 autocorrelation. This expression is used only as a descriptive AR(1) approximation. In particular, negative autocorrelation can produce an estimated $N_{\mathrm{eff}}>N$; such a value should not be interpreted as a literal increase in the number of independent observations.

\begin{table}[!ht]
	\centering
	\caption{Temporal-structure diagnostics for semantic relevance and
		surprisal. Alice values were computed from its single word sequence.
		Narratives values are unweighted means across the four stories. Word-level
		spectral centroids are expressed in cycles per word, whereas HRF-convolved
		centroids are expressed in cycles per TR. Narratives
		$N_{\mathrm{eff}}$ values are mean within-story diagnostics and are not
		directly comparable with the Alice values.}
	\label{tab:semrel-smoothness-diagnostics}
	\scalebox{0.78}{
		\begin{tabular}{lllrrrr}
			\toprule
			Dataset & Representation & Predictor
			& Lag-1 ACF & $N_{\mathrm{eff}}$
			& Low-frequency power & Spectral centroid \\
			\midrule
			Alice
			& Word level
			& Semantic-similarity metric
			& 0.485 & 740.3 & 0.275 & 0.149 \\
			
			Alice
			& Word level
			& Surprisal
			& 0.031 & 2001.9 & 0.145 & 0.241 \\
			
			Alice
			& Word level
			& Smoothed surprisal
			& 0.486 & 737.5 & 0.324 & 0.145 \\
			\midrule
			
			Narratives
			& Word level
			& Semantic relevance
			& $-0.148$ & 1994.2 & 0.063 & 0.279 \\
			
			Narratives
			& Word level
			& Surprisal
			& 0.055 & 1362.2 & 0.139 & 0.237 \\
			\midrule
			
			Narratives
			& HRF-convolved TR level
			& Semantic relevance
			& 0.889 & 27.4 & 0.521 & 0.061 \\
			
			Narratives
			& HRF-convolved TR level
			& Surprisal
			& 0.907 & 22.6 & 0.558 & 0.054 \\
			\bottomrule
	\end{tabular}}
\end{table}

The diagnostics reveal different patterns across the two datasets. In Alice, the semantic-similarity metric was substantially more autocorrelated than the original surprisal series, contained more low-frequency power, and had a lower spectral centroid. Thus, temporal smoothness remains a relevant alternative explanation for the Alice comparison. Smoothing surprisal successfully matched its lag-1 autocorrelation to that of the semantic-similarity metric, providing a dataset-specific sensitivity analysis.

The same pattern was not observed in Narratives. At the word level, semantic relevance showed a negative mean lag-1 autocorrelation, less low-frequency power, and a higher spectral centroid than surprisal. It was therefore not the smoother of the two original Narratives predictors. After canonical-HRF convolution and TR sampling, both predictors became strongly autocorrelated, as expected, but surprisal was slightly more autocorrelated and contained slightly more low-frequency power than semantic relevance. Consequently, the Narratives semantic-relevance findings cannot be explained by semantic relevance having a generally smoother temporal spectrum than surprisal.

These diagnostics do not rule out all forms of temporal confounding. HRF convolution induces strong autocorrelation in every word-derived predictor, and residual BOLD autocorrelation can still inflate precision if it is not represented in the statistical model. The temporal-spectrum results therefore address the specific claim that semantic relevance was intrinsically smoother than surprisal, whereas participant-level inference, explicit residual autocorrelation modeling, and appropriately structured temporal null models address separate aspects of temporal dependence.

\subsection{Frequency matching and its interpretation}

In the previous dataset, surprisal was smoothed to match the lag-1 autocorrelation of semantic relevance. That comparison is not directly applicable to the current Narratives analysis. In Narratives, the original word-level semantic-relevance series was not smoother than surprisal: semantic relevance showed a negative mean lag-1 autocorrelation and less low-frequency power. After canonical-HRF convolution, both predictors became strongly autocorrelated, but surprisal was slightly more autocorrelated and contained slightly more low-frequency power than semantic relevance.

Consequently, smoothing Narratives surprisal to match semantic relevance would neither reproduce the original rationale for the control nor provide a conservative test. It would further smooth the predictor that was already at least as smooth as semantic relevance in the representation entered into the BOLD model. We therefore did not use an autocorrelation-matched surprisal analysis as a Narratives inferential test. The Alice smoothed-surprisal series reported in Table~\ref{tab:semrel-smoothness-diagnostics} is retained only as a dataset-specific temporal diagnostic and should not be interpreted as a Narratives control.

These results directly weaken the simplest smoothness-only account of the Narratives findings. Semantic relevance could not have obtained a general advantage merely by possessing greater lag-1 autocorrelation or more low-frequency power than surprisal, because it did not possess either property in the Narratives predictors. This comparison does not, however, eliminate other forms of temporal confounding, including shared alignment with story structure, HRF-induced autocorrelation, and residual dependence in the BOLD signal.

\subsection{Scope of the temporal-null evidence}

The present temporal-robustness evidence instead comes from three sources. First, the raw and HRF-convolved predictor diagnostics show that Narratives semantic relevance was not temporally smoother than surprisal. Second, participant-level FIR analyses reTduce the risk that significance was driven by treating repeated TR-level observations as independent. Residual temporal dependence remains a limitation of the
transformed-BOLD GAMMs and may affect the precision of the
smooth-term tests. %Third, \textbf{the sensitivity of the GAMM results} to AR(1) modeling demonstrates that residual temporal dependence affects the apparent precision of transformed-BOLD associations and must be considered when interpreting widespread smooth-term significance.

These analyses constrain, but do not completely exclude, a temporal-structure explanation. The predictor diagnostics address differences in autocorrelation and spectral power, whereas participant-level inference addresses generalization across listeners. Neither analysis constitutes a direct temporal-spectrum-matched null test of semantic specificity. A stronger confirmatory analysis would phase-randomize each Narratives predictor separately within story, preserve its Fourier amplitude spectrum, reconstruct all lagged or HRF-convolved predictors, and repeat the complete 47-participant, 20-ROI analysis using a sufficiently large null distribution. Such a test should use a prespecified participant-level or held-out-story statistic and correct across the complete ROI family.

Accordingly, the current results do not support the claim that the Narratives semantic-relevance effects are uniquely semantic after comparison with all possible temporally structured null predictors. They support the narrower conclusion that the effects cannot be attributed simply to semantic relevance having greater lag-1 autocorrelation or more low-frequency power than surprisal.

\section{Other Results}
\label{otherresults}

\begin{figure*}[t]
	\centering
	\includegraphics[width=\textwidth]{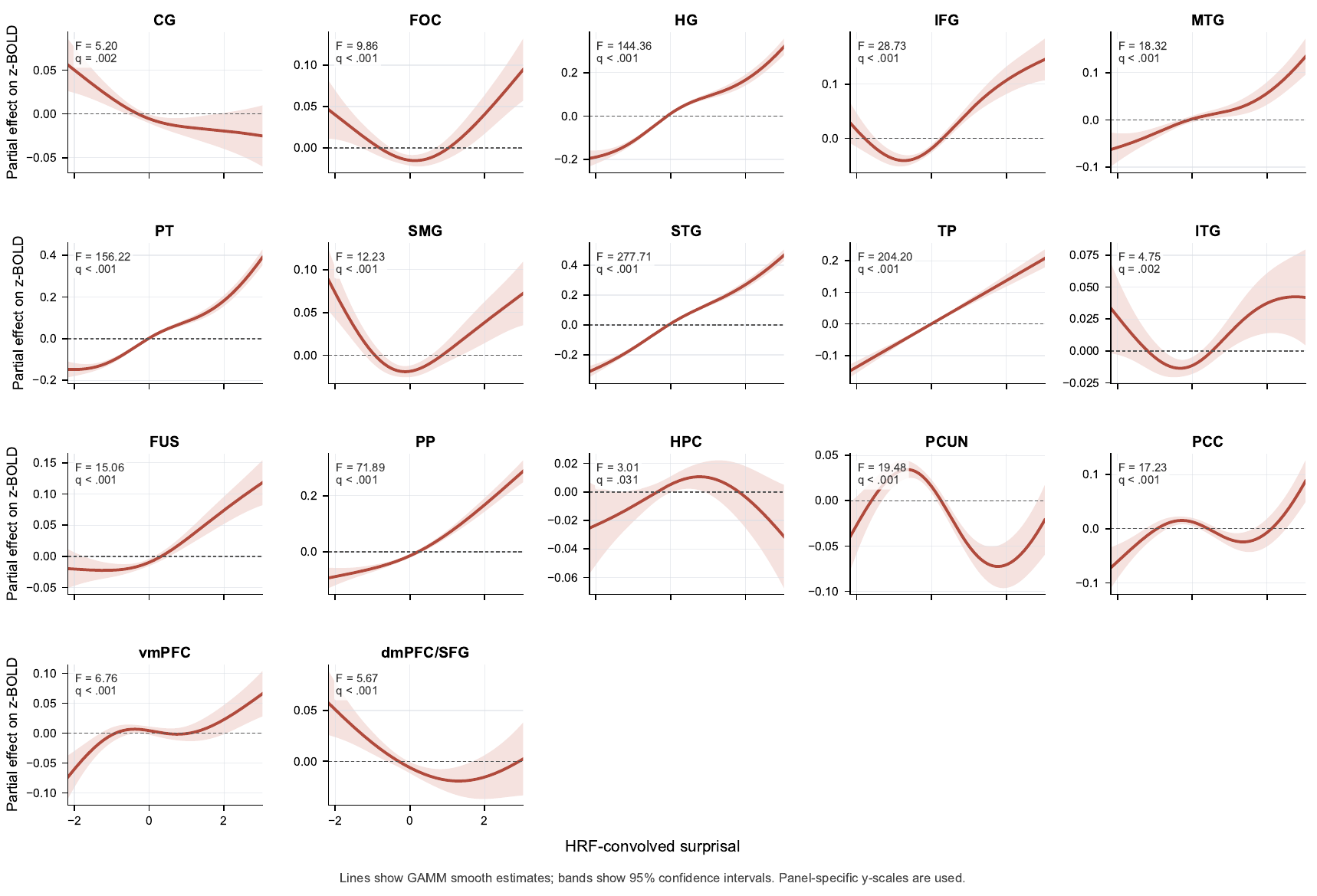}
	\caption{ GAMM partial effects for surprisal for Narratives. Curves show conditional smooth effects with the same controls as Figure~\ref{fig:narratives_sem_gamm}, including semantic relevance. Panel labels report the smooth-term $F$ statistic.}
	\label{fig:narratives_surp_gamm}
\end{figure*}

The Narratives HRF-convolved GAMM analysis showed significant partial effects of surprisal in five ROIs when semantic relevance was modeled in the same candidate-pair specification. Across
these regions, the fitted smooths showed a broadly similar near-linear positive pattern: transformed BOLD was lower at low surprisal values and higher at moderate-to-high surprisal values. Thus, surprisal showed limited but detectable associations with transformed BOLD in this GAMM framework, although this pattern contrasts with the FIR/deconvolution analyses of original BOLD, where surprisal did not show robust HRF-weighted effects. The partial effect of surprisal in the Narrative dataset is illustrated in Figure~\ref{fig:narratives_surp_gamm}.
%The Moth GAMM analysis showed significant nonlinear partial effects of surprisal in the 5 ROIs. Across these regions, the fitted curves showed a broadly similar pattern: transformed BOLD tended to be lower at low surprisal values and higher at moderate-to-high surprisal values, with some flattening at the upper end of the surprisal range. Thus, surprisal showed detectable nonlinear associations with transformed BOLD in the GAMM framework, although this pattern contrasts with the FIR/deconvolution analyses of original BOLD, where surprisal did not show robust delayed HRF-weighted effects. The partial effect of surprisal is illustrated in Figure~\ref{fig:surprisal}.

\begin{figure}[!ht]
	\centering
	\includegraphics[width=0.92\textwidth]
	{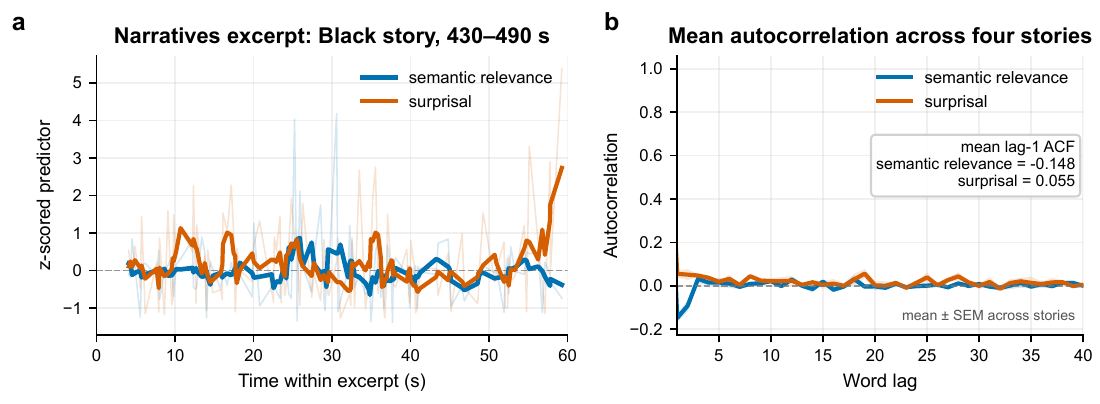}
	
	\caption{\textbf{Temporal structure of semantic relevance and surprisal in the Narratives dataset.}
		(\textbf{a}) Representative 60-s excerpt from the \textit{Black} story
		(430--490~s). Faint lines show the z-scored word-level predictor values,
		and thick lines show centered five-word rolling means included for
		visualization. The excerpt was selected for dense word-onset coverage
		rather than on the basis of either predictor's values. The horizontal axis
		represents time relative to the beginning of the excerpt, calculated from
		the true word onset times.
		(\textbf{b}) Word-lag autocorrelation functions averaged across the four
		Narratives stories; shaded bands indicate $\pm 1$ SEM across stories.
		Semantic relevance showed a negative mean lag-1 autocorrelation
		($r_1=-0.148$), whereas surprisal showed a small positive lag-1
		autocorrelation ($r_1=0.055$). Autocorrelations were close to zero at most
		subsequent lags. Therefore, semantic relevance was not more temporally
		autocorrelated than surprisal in the original Narratives word sequences.
		These word-level diagnostics do not represent the additional temporal
		smoothness introduced when predictors are convolved with the canonical
		HRF.}
	\label{fig:smoothness}
\end{figure}

Further, the temporal-structure analysis showed that semantic relevance and surprisal differed at short word lags in the Narratives dataset. In the representative 60-s excerpt, both predictors exhibited substantial word-to-word variation, and their locally averaged trajectories varied over time. Across the four stories. However, semantic relevance had a negative mean lag-1 autocorrelation, whereas surprisal had a small positive lag-1 autocorrelation ($-0.148$ vs.\ $0.055$). The autocorrelations of both predictors approached zero at subsequent word lags. In this way, the original Narratives semantic-relevance series was not temporally smoother than surprisal at the word level. This finding argues against the simple explanation that semantic relevance produced stronger BOLD associations merely because it had greater short-range autocorrelation. It does not eliminate temporal dependence introduced by HRF convolution or remaining autocorrelation in the BOLD response. The result is summarized in Figure~\ref{fig:smoothness}.

\begin{figure}[!ht]
	\centering
	\includegraphics[width=\textwidth]
	{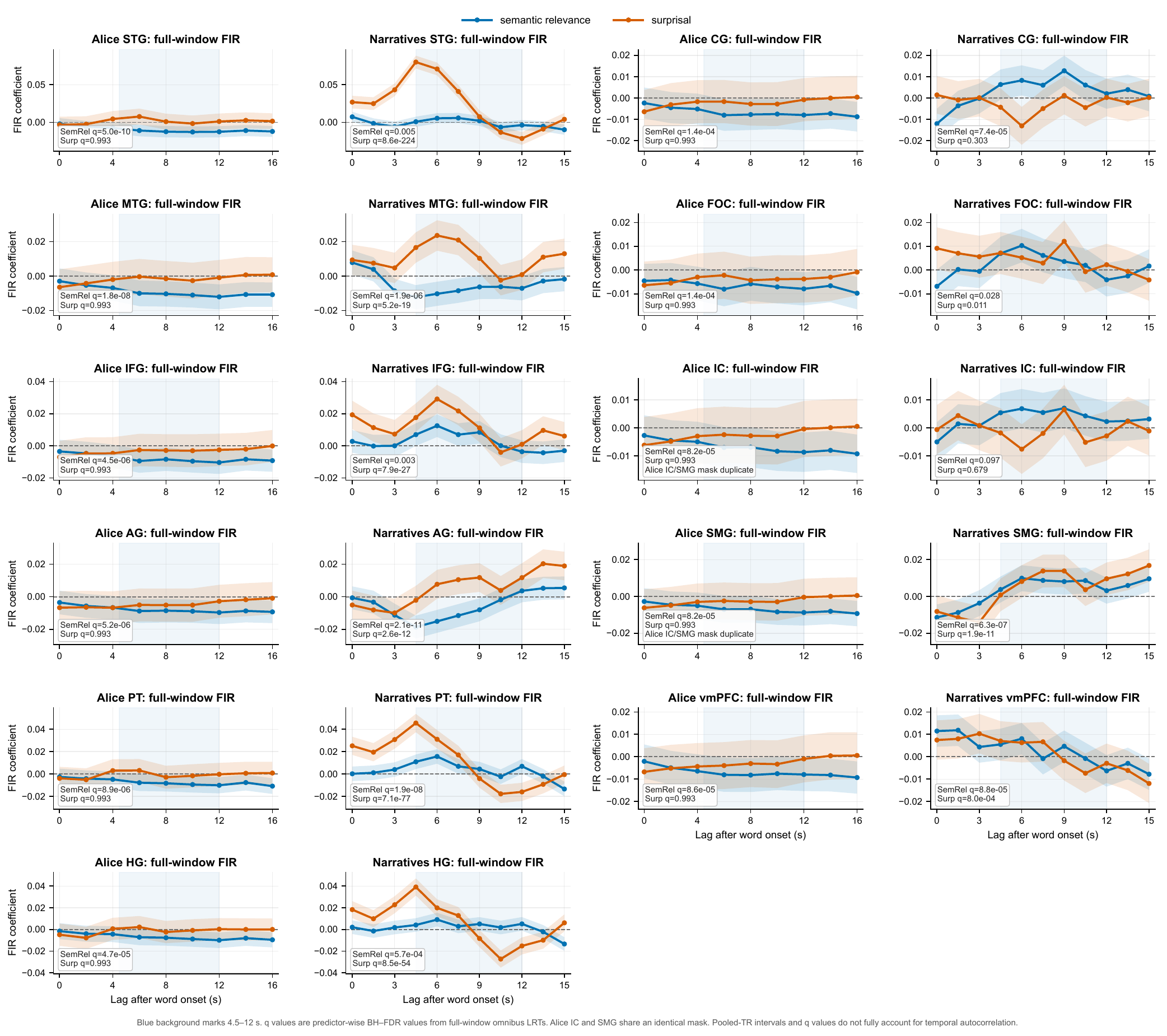}
	
	\caption{\textbf{Full-window FIR profiles in ROIs shared by Alice and Narratives.}
		Each Alice--Narratives panel pair presents the conditional FIR coefficients
		for semantic relevance (blue) and surprisal (orange) in one of the 11
		common ROI labels. Points and lines show coefficient estimates across lags
		after word onset; shaded bands show nominal 95\% confidence intervals.
		Within each ROI pair, the vertical-axis limits are matched across datasets
		to facilitate visual comparison. The light-blue background marks the
		prespecified 4.5--12-s hemodynamic window for reference, although the
		displayed $q$ values correspond to full-window omnibus target-deletion
		likelihood-ratio tests: 0--16~s in Alice and 0--15~s in Narratives.
		BH--FDR correction was performed separately for semantic relevance and
		surprisal across the complete dataset-specific ROI families (12 ROIs for
		Alice and 20 ROIs for Narratives), rather than only across the ROIs shown
		here. Semantic relevance survived correction in all 11 common Alice ROI
		labels and in 10/11 Narratives ROIs; surprisal survived in 0/11 and 9/11,
		respectively. The omnibus $q$ values are directionless and therefore do
		not test the sign of any individual lag coefficient. Alice IC and SMG
		were derived from an identical mask and should not be interpreted as
		independent anatomical observations. The plotted intervals and pooled
		TR-level omnibus $q$ values do not fully account for within-participant
		temporal autocorrelation.}
	\label{fig:common-roi-fir}
\end{figure}

To compare the temporal response profiles across datasets, we examined the 11 ROI labels available in both Alice and Narratives. In Alice, the full-window omnibus effect of semantic relevance survived BH--FDR correction in all 11 common ROI labels, whereas surprisal did not survive correction in any of them. The Alice semantic-relevance coefficients were predominantly negative at delayed lags. In Narratives, semantic relevance survived correction in 10/11 common ROIs, with IC as the exception, whereas surprisal survived in 9/11 ROIs. The Narratives profiles were more heterogeneous in shape and direction: several regions showed biphasic or lag-dependent responses, and surprisal produced substantial delayed effects in multiple temporal, frontal, and parietal ROIs. Thus, the cross-dataset comparison supports a broadly distributed association between semantic relevance and BOLD, but it does not show a uniform response direction or a consistent semantic-relevance advantage over surprisal. The profiles are summarized in Figure~\ref{fig:common-roi-fir}.

\section{Supplementary Descriptive Connectivity Analyses}
\label{connectivity}

To provide network-level context for the ROI analyses, we visualized pairwise BOLD correlations among selected Narratives ROIs during naturalistic listening. Correlations were first calculated separately within each run. The resulting coefficients were Fisher-$z$ transformed and averaged across runs within each participant, after which participants were weighted equally in the group estimate. The analysis included all 47 participants and 185 runs. Only ROI pairs with a group-mean correlation of $r\geq .40$ are displayed.

These analyses are descriptive. The threshold was used to control the visual density of the network diagrams and does not represent a statistical significance threshold. The connectivity results were not used to test semantic relevance or surprisal and cannot establish that either predictor modulated communication between regions. Rather, they illustrate the background covariance structure within which the ROI-level GAMM and FIR/deconvolution associations were estimated.

\begin{figure}[!ht]
	\centering
	\includegraphics[width=0.96\textwidth]
	{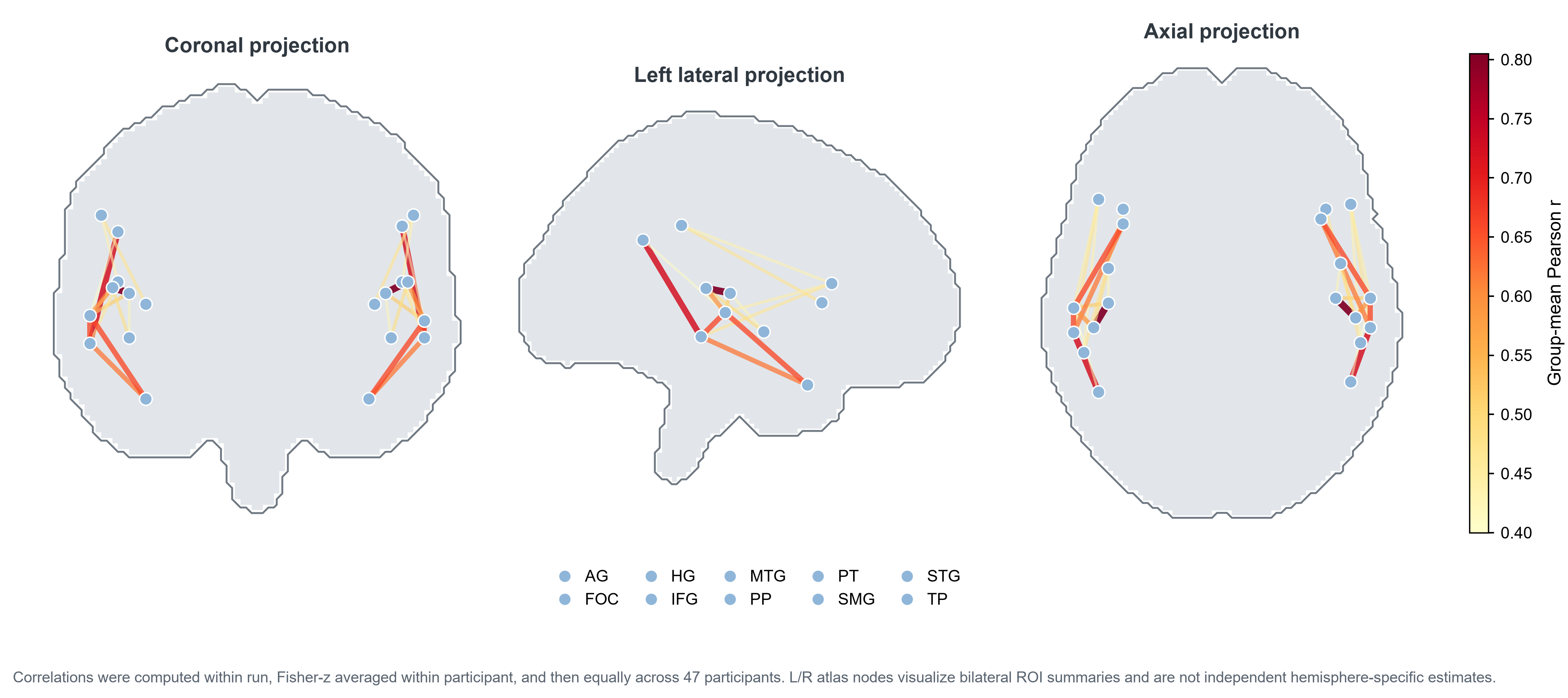}
	
	\caption{\textbf{Descriptive language-network BOLD connectivity in Narratives.}
		Nodes represent 10 selected bilateral Harvard--Oxford ROI summaries:
		AG, FOC, HG, IFG, MTG, PP, PT, SMG, STG, and TP. Edges represent the 14
		unique ROI pairs with participant-equal group-mean Pearson correlations
		of $r\geq .40$; edge color and width encode correlation magnitude.
		Correlations were calculated within run, Fisher-$z$ averaged across runs
		within participant, and then averaged with equal weighting across all 47
		participants. The same bilateral ROI summaries are projected onto
		left- and right-sided atlas locations for visualization; the displayed
		hemispheric nodes are therefore not independent hemisphere-specific
		estimates. The $r=.40$ threshold is a visualization criterion rather than
		an inferential significance threshold, and the figure does not test
		semantic-relevance or surprisal effects.}
	\label{fig:appendix-language-network}
\end{figure}

The language-network visualization in Figure~\ref{fig:appendix-language-network} contained 14 unique ROI pairs above the display threshold. The strongest coupling was observed between HG and PT ($r=.805$), followed by AG--MTG ($r=.726$), MTG--STG ($r=.661$), and STG--TP ($r=.658$). Additional suprathreshold relationships linked temporal and auditory regions with inferior-frontal, parietal, and supramarginal ROIs. This pattern is consistent with distributed BOLD covariation during continuous speech comprehension, but it should not be interpreted as evidence that the displayed regions constitute a uniquely defined or predictor-specific functional network.

\begin{figure}[!ht]
	\centering
	\includegraphics[width=0.96\textwidth]
	{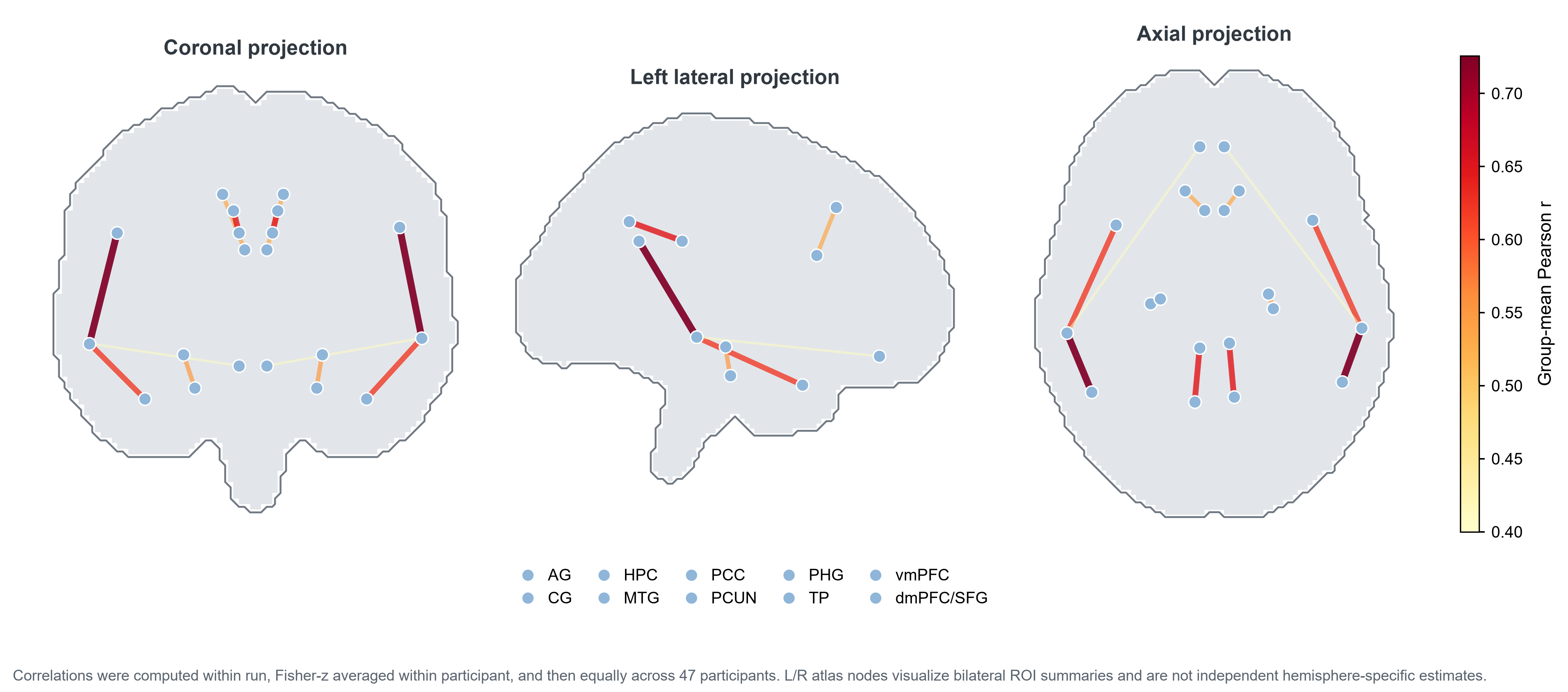}
	
	\caption{\textbf{Descriptive memory/narrative-network BOLD connectivity in Narratives.}
		Nodes represent selected bilateral Harvard--Oxford ROI summaries associated
		with memory, contextual integration, or narrative processing: AG, CG, HPC,
		MTG, PCC, PCUN, PHG, TP, vmPFC, and dmPFC/SFG. Edges represent the six
		unique ROI pairs with participant-equal group-mean Pearson correlations
		of $r\geq .40$; edge color and width indicate correlation magnitude.
		Correlations were computed within run, Fisher-$z$ averaged within
		participant, and then averaged equally across 47 participants. Bilateral
		ROI summaries are displayed at left- and right-sided atlas locations only
		to provide anatomical projections and should not be interpreted as
		independent hemisphere-specific estimates. The visualization describes
		background ROI covariance and is not a statistical test of semantic
		relevance, surprisal, effective connectivity, or directional information
		flow.}
	\label{fig:appendix-memory-network}
\end{figure}

The complementary memory/narrative visualization contained six unique ROI pairs above the same threshold (Figure~\ref{fig:appendix-memory-network}). These included AG--MTG ($r=.726$), PCC--PCUN ($r=.646$), MTG--TP ($r=.621$), HPC--PHG ($r=.551$), CG--dmPFC/SFG ($r=.536$), and MTG--vmPFC ($r=.409$). The pattern demonstrates appreciable covariation among regions commonly discussed in relation to episodic memory, internally directed processing, discourse context, and narrative comprehension. Nevertheless, these anatomical associations do not establish that semantic relevance specifically recruited a memory or situation-model network. They provide only descriptive context for interpreting semantic-relevance effects observed in some of the same regions.

Overall, the two visualizations show that the analyzed ROIs exhibited substantial shared BOLD dynamics during narrative listening. This dependence reinforces the need to interpret ROI-level predictor effects as associations estimated within a correlated distributed system, rather than as independent activations of isolated regions. As zero-lag BOLD correlations can reflect shared stimulus timing, common physiological fluctuations, preprocessing, and indirect network relationships, these descriptive analyses do not support causal or directional conclusions.

\end{document}